%% file: main.tex
\documentclass{article}

\usepackage{microtype}
\usepackage{graphicx}
\graphicspath{ {./figures/} }
\usepackage{subfigure}
\usepackage{booktabs} % for professional tables
\usepackage{caption}
\usepackage{fancyvrb}

\usepackage{bbm}

\usepackage[accepted]{icml2023}

\usepackage{paralist}

\usepackage{amsfonts}
\usepackage{amsmath}
\usepackage{amssymb}
\usepackage{amsthm}

\usepackage{hyperref}

\usepackage{parskip}

\usepackage{enumitem}

\usepackage[english]{babel}

\theoremstyle{plain}
\newtheorem{theorem}{Theorem}[section]

\theoremstyle{definition}
\newtheorem{definition}[theorem]{Definition}

\newtheorem*{example}{Example}
\theoremstyle{remark}

\newcommand{\cosp}[1]{\cos \left ( #1 \right )}
\newcommand{\sinp}[1]{\sin \left ( #1 \right )}

\newcommand{\R}{\mathbb{R}}

\newcommand{\gcalgorithm}{group composition via representations}
\newcommand{\abbrevalgorithm}{GCR}

\DeclareMathOperator{\tr}{tr}
\DeclareMathOperator*{\argmax}{argmax}

\icmltitlerunning{A Toy Model of Universality: Reverse Engineering how Networks Learn Group Operations}
\begin{document}

\twocolumn[
%\icmltitle{Universality in Neural Networks: Evidence from Group Composition Tasks via Mathematical Representation Theory}
\icmltitle{A Toy Model of Universality:\\ Reverse Engineering how Networks Learn Group Operations}



\icmlsetsymbol{equal}{*}

\begin{icmlauthorlist}
\icmlauthor{Bilal Chughtai}{indep}
\icmlauthor{Lawrence Chan}{berk}
\icmlauthor{Neel Nanda}{indep}
\end{icmlauthorlist}

\icmlaffiliation{indep}{Independent}
\icmlaffiliation{berk}{UC Berkeley}

\icmlcorrespondingauthor{Bilal Chughtai}{brchughtaii@gmail.com}

\icmlkeywords{Machine Learning, Mechanistic Interpretability, Interpretability, Representation Theory, Circuits, Universality, Convergent Learning, Group Theory, ICML}

\vskip 0.3in]



\printAffiliationsAndNotice{}  % leave blank if no need to mention equal contribution

\begin{abstract}

Universality is a key hypothesis in mechanistic interpretability -- that different models learn similar features and circuits when trained on similar tasks. In this work, we study the universality hypothesis by examining how small neural networks learn to implement group composition. We present a novel algorithm by which neural networks may implement composition for any finite group via mathematical representation theory. We then show that networks consistently learn this algorithm by reverse engineering model logits and weights, and confirm our understanding using ablations. By studying networks of differing architectures trained on various groups, we find mixed evidence for universality: using our algorithm, we can completely characterize the family of circuits and features that networks learn on this task, but for a given network the precise circuits learned -- as well as the order they develop -- are arbitrary.

\end{abstract}

\section{Introduction}

\begin{figure}
    \centering
    \vspace{15pt}
    \includegraphics[width=0.9\columnwidth]{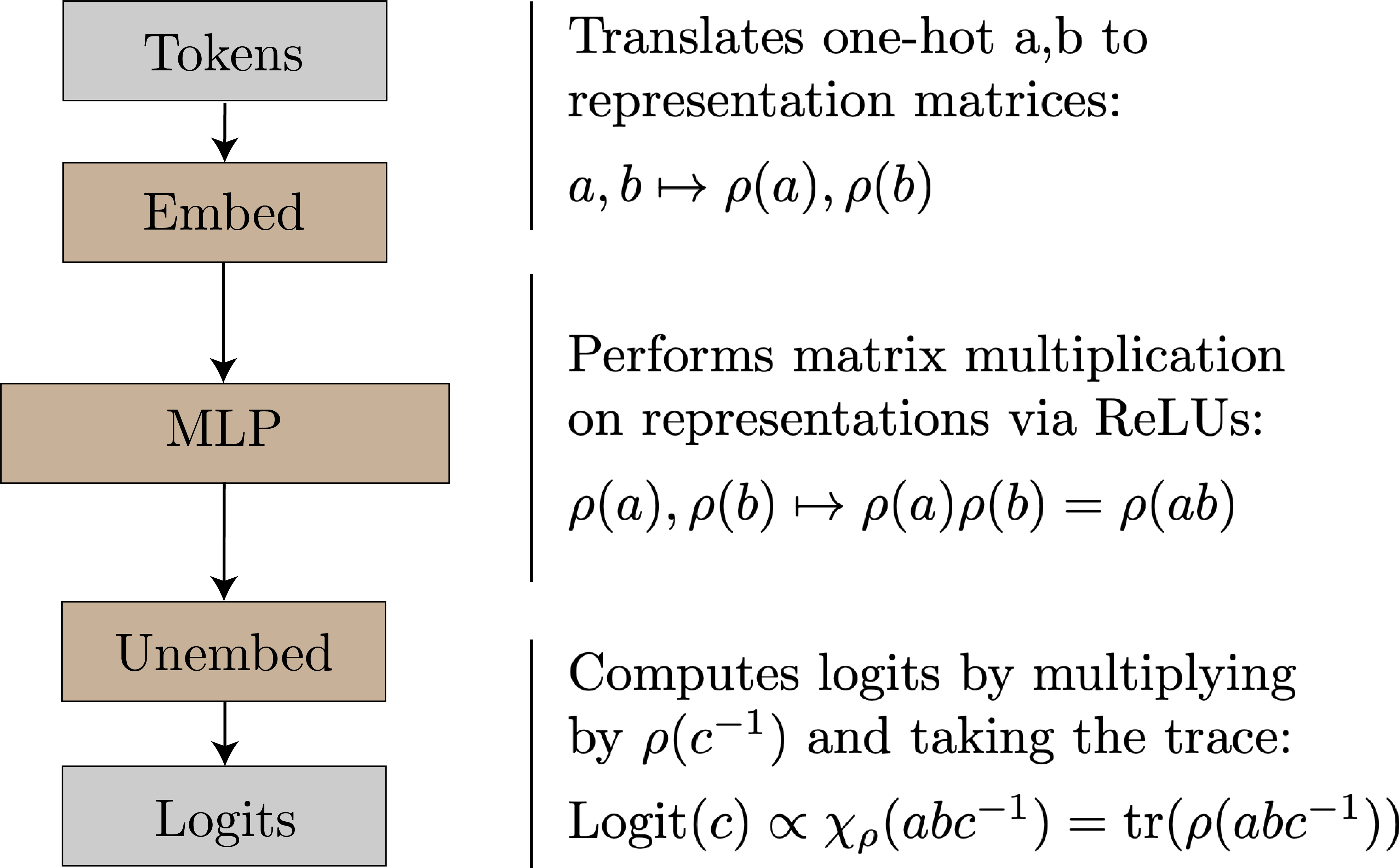}
    \caption{The algorithm implemented by a one hidden layer MLP for arbitrary group composition. Given two input group elements $a$ and $b$, the model learns representation matrices $\rho(a)$ and $\rho(b)$ in its embeddings. Using the ReLU activations in its MLP layer, it then multiplies these matrices, computing $\rho(a)\rho(b) = \rho(ab)$. Finally, it `reads off' the logits for each output group element $c$ by computing \textit{characters} -- the matrix trace $\tr\rho(abc^{-1})$, denoted $\chi_\rho(abc^{-1})$, which is maximized when $c=ab$.}
    \label{fig:architecture}
    % \vspace{-0.5cm}
\end{figure}

Do models converge on the same solutions to a task, or are the algorithms implemented arbitrary and unpredictable? The \textit{universality hypothesis} \cite{olahZoomIntroductionCircuits2020, liConvergentLearningDifferent2016} asserts that models learn similar features and circuits across different models when trained on similar tasks. This is an open question of significant importance to the field of \textit{mechanistic interpretability}. The field focuses on reverse engineering state-of-the-art models by identifying circuits \cite{elhageMathematicalFrameworkTransformer21, olssonIncontextLearningInduction2022, nandaProgressMeasuresGrokking2023, wangInterpretabilityWildCircuit2022}, subgraphs of networks consisting sets of tightly linked features and the weights between them.\cite{olahZoomIntroductionCircuits2020}. %Under the strongest form of universality, the study of universality in neural networks may give insights relevant to biological brains - features being interpretable to \textit{us} demonstrates that at least one class of biological neural network can too learn these features. 
Recently, the field of mechanistic interpretability has increasingly shifted towards finding small, toy models easier to interpret, and employing labor intensive approaches to reverse-engineering specific features and circuits in detail \cite{elhageMathematicalFrameworkTransformer21, wangInterpretabilityWildCircuit2022, nandaProgressMeasuresGrokking2023}. If the universality hypothesis holds, then the insights and principles found by studying small models will transfer to state-of-the-art models that are used in practice. But if universality is false, then although we may learn some general principles from small models, we should shift focus to developing scalable, more automated interpretability techniques that can directly interpret models of genuine interest.%it may suggest that the field's work is altogether intractable.

In this work, we study to what extent the universality hypothesis is true by interpreting networks trained on composition of group elements in various finite groups  \footnote{Code and a demo notebook are available at \href{https://github.com/bilal-chughtai/rep-theory-mech-interp}{https://github.com/bilal-chughtai/rep-theory-mech-interp}}. We focus on composition of arbitrary groups as this defines a large family of related tasks, forming an algorithmic test bed for investigating universality. We first exhibit a general algorithm by which networks can compute compositions of elements in an arbitrary finite group, using concepts from the mathematical field of representation\footnote{We note our use of the word `representation' is distinct to the usual use of the word representation in the ML literature.} and character theory. We do this by building upon the work of \citet{nandaProgressMeasuresGrokking2023}, that reverse-engineered networks trained to grok modular addition$\pmod p$ and found the networks used a Fourier transform and trigonometry (trig) identity based algorithm to compute logits. We show that this ad-hoc, trig identity-based algorithm is a special case of our algorithm and that distinct Fourier modes are better thought of as distinct \textit{irreducible representations} of the cyclic group. Our algorithm and how we find it implemented in network components is described in Figure~\ref{fig:architecture}.

Representation theory bridges linear algebra and group theory, and studies how group elements can be thought of as matrices. At a high level, our algorithm embeds group elements as such matrices, uses its ReLU activations to perform matrix multiplication, and uses the unembed to convert back to group elements. We prove correctness of our algorithm using results from representation theory in Section~\ref{algorithm}. 

% Representation theory is of universal importance to science. It appears in chemistry in the study of the symmetries of molecules in space, and is central to quantum physics - the representation theory of certain Lie groups that encode symmetries of particular differential equations are directly responsible for the particles we predict and see in the universe. - currently in appendix

We verify our understanding of a model trained to perform group composition with four lines of evidence in Section~\ref{sec:reverse_engineer}. (1) the logits are as predicted by the algorithm over a set of key representations $\rho$. (2) the embeddings and unembeddings purely consist of a memorized lookup table, converting the inputs and outputs to the relevant representation matrices $\rho(a)$, $\rho(b)$ and $\rho(c^{-1})$. (3) the MLP neurons calculate $\rho(ab)$, and we can explicitly extract these representation matrices from network activations. Further, we can read off the neuron-logit map directly from weights, and neurons cluster by representation. (4) ablating the components of weights and activations predicted by our algorithm destroys performance, while ablating parts we predict are noise does not affect loss, and often \textit{improves} it.

Finally, we use our mechanistic understanding of models to investigate the universality hypothesis in Section~\ref{sec:universality}. We break universality down into strong and weak forms. Strong universality claims that the same features and circuits arise in all models that are trained in similar ways; weak universality claims that there are underlying principles to be understood, but that any given model will implement these principles in features and circuits in a somewhat arbitrary way. While models consistently implement our algorithm across groups and architectures by learning representation-theoretic features and circuits, we find that the choice of specific representations used by networks varies considerably. Moreover, the number of representations learned and order of representations learned is not consistent across different hyperparameters or random seeds. We consider this to be compelling evidence for weak universality, but against strong universality: interpreting a single network is insufficient for understanding behavior across networks. 

%Following \citet{nandaProgressMeasuresGrokking2023}, we too define progress measures, and replicate their results in using these to understand grokking generalisation.

\begin{figure}
    \centering
    \begin{subfigure}{}
        \includegraphics[width=0.47\columnwidth]{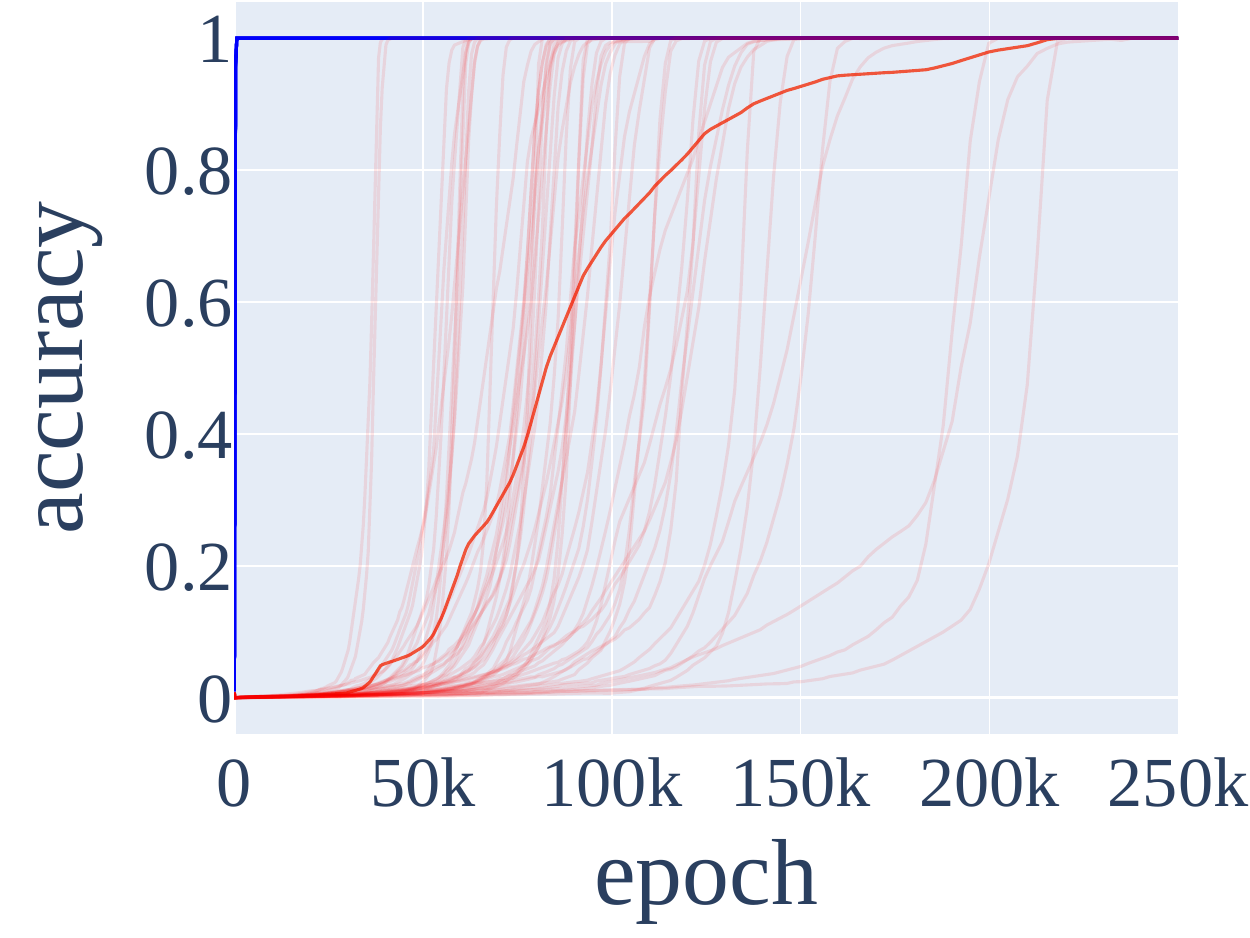}
    \end{subfigure}
    \hfill
    \begin{subfigure}{}
        \includegraphics[width=0.47\columnwidth]{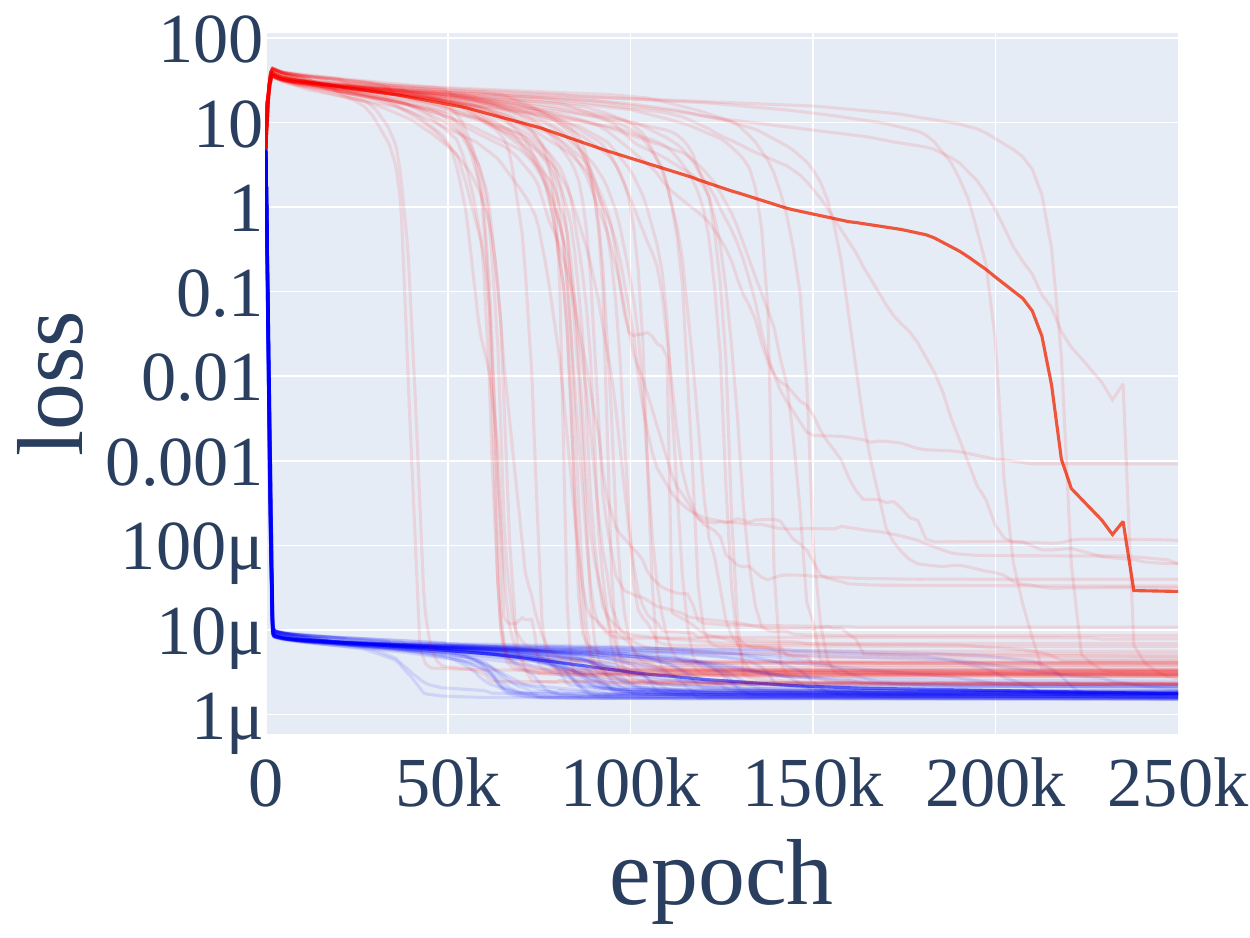}
    \end{subfigure}
    \caption{Train (blue) and test (red) accuracy (\textbf{left}) and train and test loss (\textbf{right}) of an MLP trained on group composition on $S_5$, the permutation group of order 5, over 50 random seeds. These models consistently exhibit grokking: they quickly overfit early in training, but then suddenly generalize much later. The bolded line denotes average accuracy/loss.}
    \label{fig:grokking}
    \vspace{-0.5cm}
\end{figure}

\section{Related Work}

\textbf{Comparing Neural Representations}. In the past several years, a wide variety of post-hoc approaches have been used to study the relationship between the representations learned by neural networks, initiated by \citet{liConvergentLearningDifferent2016}. Methods often compare internal representations of one network to another, though it is unclear whether these methods truly measure what we want, as networks are highly non linear and may learn similar features in different ways. Empirically however, techniques such as Canonical Correlation Analysis \cite{morcosInsightsRepresentationalSimilarity2018}, Centered Kernel Alignment \cite{kornblithSimilarityNeuralNetwork2019a} and variations are able to quantify representation similarity. Other techniques used include model stitching \cite{bansalRevisitingModelStitching2021} and neuroscience-inspired methods \cite{mehrerIndividualDifferencesDeep2020}.

\textbf{Mechanistic Interpretability and Universality.} 
In contrast, we are able to compare the learned representations of models to a known ground truth, through first reverse engineering the employed algorithm completely and thereby understanding the full set of features. We employ a Circuits-based mechanistic interpretability approach, as pioneered by \citet{cammarataThreadCircuits2020}, \citet{elhageMathematicalFrameworkTransformer21} and \citet{olssonIncontextLearningInduction2022}. In mechanistic interpretability, neural representation similarity is studied together with algorithm similarity under the term `universality'. \citet{olahZoomIntroductionCircuits2020} demonstrated the universality hypothesis in image models through the presence of curve detector and high-low frequency detector features in early layers of many models, while also showing the circuits implementing them are analogous. 

\textbf{Group Theory.} Group theoretic tasks have in the past been used to probe the capability of neural networks to perform symbolic and algorithmic reasoning. \cite{zhangUnveilingTransformersLEGO2022} evaluate and fine tune language models to implement group actions in context. \citet{liuTransformersLearnShortcuts2022} study how Transformers learn group theoretic automata. 

\textbf{Phase Changes and Emergence.} Recent work has observed emergent behavior in neural networks: models often quickly develop qualitatively different behavior as they are scaled up \cite{ganguliPredictabilitySurpriseLarge2022,weiEmergentAbilitiesLarge2022}. \citet{brownLanguageModelsAre2020} find that, while total loss scales predictably with model size, models’ ability to perform specific tasks can change abruptly with scale. %One notable example is addition, an unexpected capability for a language model to learn, and an example of a group operation. 
\citet{mcgrathAcquisitionChessKnowledge2022} find that AlphaZero quickly learns many human chess concepts between 10k and 30k training steps and reinvents human opening theory between 25k and 60k training steps. %\citet{olssonIncontextLearningInduction2022} demonstrate universality of `induction heads' across distinct large language models.

\textbf{Grokking.} Grokking is a form of emergence, first reported by \cite{powerGrokkingGeneralizationOverfitting2022}, who trained small networks on algorithmic tasks, finding that test accuracy often increased sharply, long after maximizing train accuracy. \citet{liuUnderstandingGrokkingEffective2022} construct further small examples of grokking, which they use to compute phase diagrams with four separate `phases' of learning. \citet{daviesUnifyingGrokkingDouble2022} unify the phenomena of grokking and double descent as instances of phenomena dependent on `pattern learning speeds'. Our findings agree with \citet{liuOmnigrokGrokkingAlgorithmic2022} in that grokking seems intrinsically linked to the relationship between performance and weight norms; and with \citet{barakHiddenProgressDeep2023} and \citet{nandaProgressMeasuresGrokking2023} in showing that the networks make continuous progress toward a generalizing algorithm, which may be tracked over training using continuous \textit{progress measures}.

\section{Setup and Background}

\subsection{Task Description}
\label{sec:task_description}

We train models to perform group composition on finite groups $G$ of order $|G| = n$.  The input to the model is an ordered pair $(a, b)$ with $a, b \in G$ and we train to predict the group element $c = ab$.  In our mainline experiment, we use an architecture consisting of left and right embeddings\footnote{We do not tie the left and right embeddings as we study non abelian groups.}, a one hidden layer MLP, and unembedding $W_U$. This architecture is presented in Figure~\ref{fig:architecture} and elaborated upon in Appendix~\ref{sec:architecture_details}. We note that the task presented in \citet{nandaProgressMeasuresGrokking2023} is a special case of our task, as addition mod $113$ is equivalent to composition for $G = C_{113}$, the cyclic group of $113$ elements. We train our models in a similar manner to \citet{nandaProgressMeasuresGrokking2023}, details may be found in Appendix~\ref{sec:architecture_details}.

\subsection{Mathematical Representation Theory}

\label{sec:rep_theory}
The core claims of our work build on a rich sub-field of pure mathematics named Representation Theory. We introduce the key definitions and results used throughout here, but discuss and motivate other relevant results in Appendix~\ref{sec:appendix_rep_theory}. Further details and proofs beyond this may be found in e.g. \citet{alperinGroupsRepresentations1995}. 

A (real) \textbf{representation} is a homomorphism, i.e. a map preserving the group structure, $\rho: G \rightarrow GL(\R^d)$ from the group $G$, to a $d$-dimensional general linear group, the set of invertible square matrices of dimension $d$. Representations are in general \textit{reducible}, in a manner we make precise in the Appendix. For each group $G$, there exist a finite set of fundamental \textbf{irreducible representations}. The \textbf{character} of a representation is the trace of the representation $\chi_\rho: G \to \R$ given by $\chi_\rho(g) = \tr(\rho(g))$. A key fact our algorithm depends on is that character's are maximal when $\rho(g) = I$, the identity matrix (Theorem~\ref{thm:trace}). In particular, the character of the identity element, $\chi_\rho (e)$, is maximal.

\begin{example}
The cyclic group $C_n$ is generated by a single element $r$ and naturally represents the set of rotational symmetries of an n-gon, where $r$ corresponds to rotation by $2\pi/n$. This motivates a 2 dimensional representation -- a set of $n$ $2\times2$ matrices, one for each group element:

%$$ \rho(r^k) = \begin{pmatrix}
%\cos\theta & -\sin\theta \\
%\sin\theta & \cos\theta
%\end{pmatrix}   $$

$$ \rho(r^k) = \begin{pmatrix}
\cosp{\frac{2\pi k}{n}} & -\sinp{\frac{2\pi k}{n}} \\
\sinp{\frac{2\pi k}{n}} & \cosp{\frac{2\pi k}{n}}
\end{pmatrix}   $$

for element $r^k$, corresponding to rotation by $\theta = 2\pi k/n$. This representation is irreducible, since there is no subspace of $\R^2$ on which the set of rotation matrices restricts -- they each rotate the whole space. The character of each representation element is the trace $\chi_\rho(r^k) = 2\cos\theta$, which is maximized at $\theta = 0$, where the group element $r^0 = e$ and corresponding matrix $I_2$ are both the identity.
\end{example}

\section{An Algorithm for Group Composition}
\label{algorithm}

We now present an algorithm, which we call \gcalgorithm{} (\abbrevalgorithm), on an arbitrary group $G$ equipped with a representation $\rho$ of dimension $d$. The algorithm and it's map onto network components are described in Figure~\ref{fig:architecture}. We are not aware of this algorithm existing in any prior literature.

\begin{enumerate}[label=(\arabic*), noitemsep]
    \item Map inputs $a$ and $b$ to $d \times d$ matrices $\rho(a)$, $\rho(b)$. %We think of each of these $d\times d$ matrices as a vector of dimension $d^2$.
    \item Compute the matrix product $\rho(a)\rho(b) = \rho(ab)$. 
    \item For each output logit $c$, compute the characters $\tr(\rho(ab)\rho(c^{-1})) =  \tr(\rho(abc^{-1})) = \chi_\rho(abc^{-1})$. 
\end{enumerate}

Crucially, Theorem~\ref{thm:trace} implies $ab \in \argmax_c \chi_\rho(abc^{-1})$, so that logits are maximised on $c^*=ab$, where $abc^{-1} = e$. If $\rho$ is \textit{faithful} (see Definition~\ref{def:faithful_rep}), this argmax is unique.

In our networks, we find the terms $\rho(a)$ and $\rho(b)$ in the embeddings and $\rho(ab)$ in MLP activations. Note, as $\rho(ab)$ is present in the final hidden layer activations and $W_U$ learns $\rho(c^{-1})$ in weights, the map to logits is entirely linear:
\begin{equation}
\rho(ab) \rightarrow \tr \rho(ab)\rho(c^{-1}) = \sum_{ij} \left(\rho(ab) \odot \rho(c^{-1})^T\right)_{ij}
\label{eq:linear_map}
\end{equation}
where $\odot$ denotes the element-wise product of matrices.

Each finite group $G$ is equipped with a finite set of $k$ \textit{irreducible} representations (Definition \ref{def:irrep}) Since any representation may be decomposed into a finite set of irreducible representations (Theorems~\ref{th:maschke} and \ref{th:sum_of_dims}) we may restrict our attention to these irreducible representations. It is then useful to think about our algorithm for a fixed group $G$ as a \textit{family} of $k$ independent circuits indexed by choice irreducible representation $\rho$. In general, a single network may choose any subset of these $k$ circuits to implement, so that the observed logits are a linear combination of characters from multiple representations. From now on, each representation may be assumed to be irreducible, and we will drop the word. Since each representation has $\chi_{\rho}(abc^{-1})$ maximized on the correct answers, using multiple representations gives constructive interference at $c^* = ab$, giving $c^*$ a large logit. Theorem~\ref{th:schur_matrix} implies characters are orthogonal over distinct representations, a fact we use in Section~\ref{sec:logit_attribution}.

\begin{example}
Our \abbrevalgorithm{} algorithm is a generalization of the seemingly ad-hoc algorithm presented in \citet{nandaProgressMeasuresGrokking2023} for modular addition, which in our framing is composition on the cyclic group of 113 elements, $C_{113}$. Each element of our algorithm maps onto their Fourier multiplication algorithm, with representations $\rho = \mathbf{2_k}$ (which we define in Appendix~\ref{sec:c_reps}) corresponding to frequency $\omega_k = \frac{2\pi k}{n}$.

\citet{nandaProgressMeasuresGrokking2023} found embeddings learn the terms $\cosp{\omega_k a}$, $\sinp{\omega_k a}$, $\cosp{\omega_k b}$ and $\sinp{\omega_k b}$, precisely the matrix elements of $\rho(a)$ and $\rho(b)$. The terms $\cosp{\omega_k(a+b)}$ and $\sinp{\omega_k(a+b)}$ found in the MLP neurons correspond directly to the matrix elements of $\rho(ab)$. Finally we find by direction calculation, or by using the group homomorphism property of representations, that the characters:
\begin{align*}
&\chi(abc^{-1}) \\
&= \tr\left(\rho(abc^{-1})\right) \\
&= \tr\begin{pmatrix}
\cosp{\omega_k(a+b-c)} & -\sinp{\omega_k(a+b-c)} \\
\sinp{\omega_k(a+b-c)} & \cosp{\omega_k(a+b-c)} 
\end{pmatrix}  \\
&= 2\cosp{\omega_k(a+b-c)}
\end{align*}
are precisely the form of logits found, which summed over many key frequencies $k$, corresponding to distinct irreducible representations. 
\end{example}

\section{Reverse Engineering Permutation Group Composition in a One Layer ReLU MLP}
\label{sec:reverse_engineer}
We follow the approach of \citet{nandaProgressMeasuresGrokking2023} in reverse engineering a single mainline model trained on a fixed group, and then showing our interpretation is robust and generic later in Section~\ref{sec:universality}, by analyzing models of different architectures trained on composition on several different groups, over different random initializations. We produce several lines of mechanistic evidence that the \abbrevalgorithm{} algorithm is being employed, mostly mirroring those in \citet{nandaProgressMeasuresGrokking2023}.

In our mainline experiment, we train the MLP architecture described in Section~\ref{sec:task_description} on the permutation (or symmetric) group of 5 elements, $S_5$, of order $|G| = n = 120$. Note that unlike $C_{113}$ studied by \citet{nandaProgressMeasuresGrokking2023}, $S_5$ is not abelian, so the composition is non-commutative. We present a detailed analysis of this case as symmetric groups are in some sense the most fundamental group, as every group is isomorphic to a subgroup of a symmetric group (Cayley's Theorem~\ref{th:cayley}). So, understanding composition on the symmetric group implies understanding, in theory, of composition on any group. The (non trivial) irreducible representations of $S_5$ are named sign, standard, standard\_sign, 5d\_a, 5d\_b, and 6d, are of dimensions $d = \{1, 4, 4, 5, 5, 6\}$ and are listed in Appendix~\ref{sec:s_reps}.

The \abbrevalgorithm{} algorithm predicts that \textbf{logits} are sums of characters. This is a strong claim, which we directly verify in a black-box manner -- we need not peer directly into network internals to check this. We do so by comparing the model's logits $l(a,b,c)$ on all input pairs $(a,b)$ and outputs $c$ with the algorithms character predictions $\chi_\rho(abc^{-1})$ for each representation $\rho$. We find the logits can be explained well with only a very sparse set of directions in logit space, corresponding to the characters of the `standard' and `sign' representations. From now on we call these two representations the \textit{key representations}.

The remainder of our approaches are white-box and involve direct access to internal model weights and activations. First, we inspect the \textbf{mechanisms} implemented in model weights. We find the embeddings and unembeddings to be memorized look up tables, converting inputs and outputs to the relevant representation element in the key representations. As the number of representations learned is low, the embedding and unembedding matrices are low rank.

We then find \textbf{MLP activations} calculate $\rho(ab)$, and are able to explicitly extract these representation matrices. Additionally, MLP neurons cluster into distinct representations, and we can read off the linear map from neurons to logits as being precisely the final step of the \abbrevalgorithm{} algorithm.

Finally, we use \textbf{ablations} to confirm our interpretation is faithful. We ablate components predicted by our algorithm to verify performance is hampered, and ablate components predicted to be noise, leaving only our algorithm, and show performance is maintained.

\subsection{Logit Attribution}
\label{sec:logit_attribution}

\textbf{Logit similarity}. %Flattening the observed logit $l(a,b,c)$ over all pairs of inputs $(a,b)$ and outputs $c$ yields a $n^3$ dimensional vector. The algorithm too produces an $n^3$ dimensional tensor $\chi_\rho(a,b,c)$ for each representation, which we flatten. 
We call the correlation between the logits $l(a,b,c)$ and characters $\chi_\rho(abc^{-1})$ the \emph{logit similarity}. We call representations with logit similarity (see Appendix~\ref{sec:gory-details}) greater than $0.005$ `key'.

\begin{figure}[t]
\begin{center}
\centerline{\includegraphics[width=0.9\columnwidth]{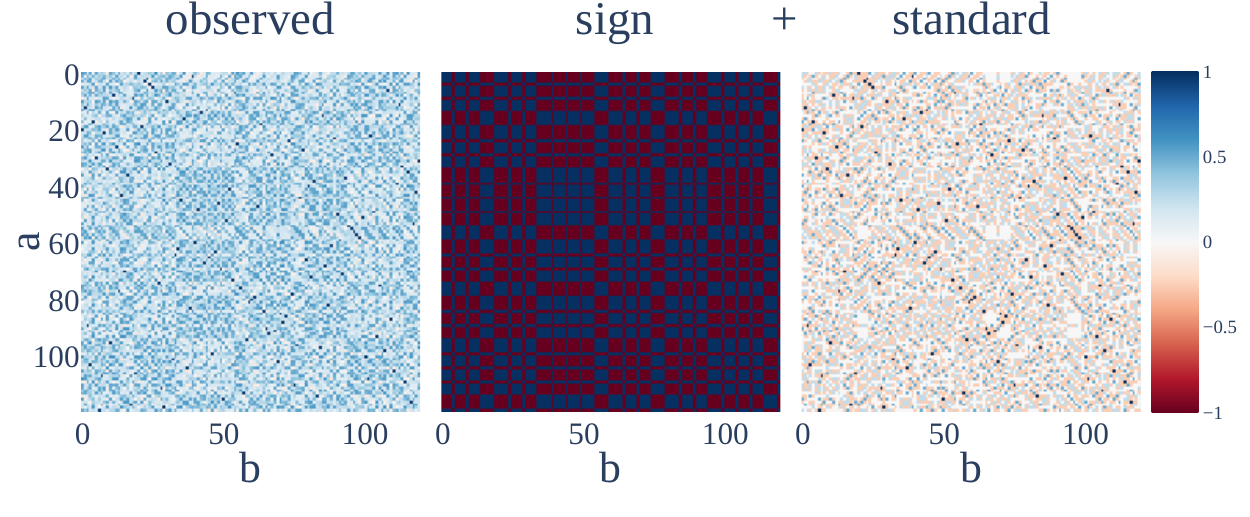}}
\caption{The observed $0^{\textrm{th}}$ logit (\textbf{left}) over all pairs of inputs $a$ (y-axis) and $b$ (x-axis). The \abbrevalgorithm{} algorithm's logit predictions $\chi_{sign}$ (\textbf{middle}) and $\chi_{standard}$ (\textbf{right}) in the key representations. The observed logit appears to be a linear combination of the characters in the key representations. Note that all logits here have been normalized to range [-1, 1].}
\label{fig:logit_cubes}
\end{center}
\vspace{-0.2cm}
\end{figure}

\begin{figure}[t]
\begin{center}
\centerline{\includegraphics[width=0.9\columnwidth]{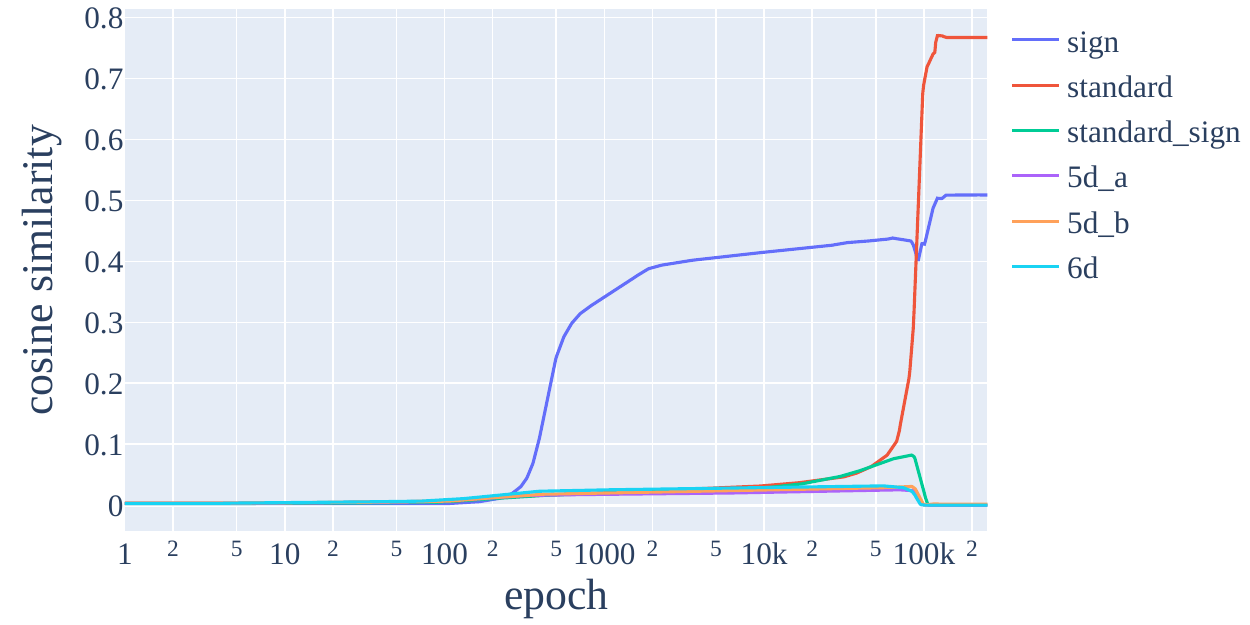}}
\vspace{-0.2cm}
\caption{Evolution of logit similarity over training for each of the six non trivial representations. We see the sign representation is learned around epoch 250, and the standard around epoch 50k. None of the other representations contribute to logits via the \abbrevalgorithm{} algorithm at the end of training. We therefore call the sign and standard representations `key'. }
\label{fig:logit_similarity}
\end{center}
\vspace{-0.5cm}
\end{figure}

Our model has logit similarity $0.509$ with $\chi_{sign}$ and $0.767$ with $\chi_{standard}$, and zero with all other representation characters. Theorem~\ref{th:schur} implies these character vectors are orthogonal, so we may approximate the logits with these two directions. Doing so explains $84.8\%$ of the variance of logits. This is surprising -- the $120$ output logits are explained well by only two directions. As confirmation for the correctness of our algorithm, if we evaluate test loss only using this logit approximation, we see a \textit{reduction} in loss by $70\%$ relatively. If we ablate the remaining $15\%$ of logits, loss does not change.

\subsection{Embeddings and Unembeddings}
\label{sec:embeddings}

Each representation is a set of $n$ $d\times d$ matrices, which by flattening we can think of as a set of $d^2$ vectors of dimension $n$. We call the subspace of $\R^n$ spanned by these vectors \textit{representation space}. Theorem~\ref{th:schur_matrix} implies these subspaces are orthogonal for distinct representations, and Theorem~\ref{th:maschke} implies the direct sum of each of these subspaces over all representations is $\R^n$. Any embedding or unembedding of $n$ group elements lies in $\R^{n \times h}$ for some $h$, so a natural operation is to project embeddings and unembeddings onto representation space over the $n$ dimension. Our definitions of embedding matrices $W_a$, $W_b$ and $W_U$ may be found in Appendix~\ref{app:mlp}, and details regarding how we perform the projection in Appendix~\ref{sec:gory-details}.

\textbf{We find evidence of representations in embeddings and unembeddings}. We find that the embedding matrices and the unembed matrix are well approximated by a sparse set of representations (Table~\ref{tb:percent_embedding_explained}), and that the representations contained in all three are the same. This is surprising: each embedding and unembedding can potentially be of rank $120$, but is only of rank $16 + 1$, corresponding precisely to the two key representations. Qualitatively, the progress of representation learning is similar across all three embedding and unembedding matrices, with each representation being learned suddenly at roughly the same time, see Figure~\ref{fig:percent_embeds}. %We discuss the learning of representations in Section~\ref{sec:rep_learning}.

\begin{table}[h]
\vspace{-0.3cm}
\caption{Percentage of embedding matrices explained by subspaces corresponding to representations. We see the same two key representations explain almost all of the variance of each embedding matrix, and the non-key representations explain almost none.}
\label{tb:percent_embedding_explained}
\vskip 0.15in
\begin{center}
\begin{small}
\begin{sc}
\begin{tabular}{lcccr}
\toprule
& $W_a$ & $W_b$ & $W_U$\\
\midrule
Sign & 6.95\% & 6.95\%& 9.58\% \\
Standard & 93.0\% & 93.0\% & 84.5\%\\
Residual & 0.00\% & 0.00\% & 5.96\%\\
\bottomrule
\end{tabular}
\end{sc}
\end{small}
\end{center}
\vskip -0.1in
\end{table}

\subsection{MLP Neurons}
\label{sec:mlp_neurons}

\textbf{MLP neurons calculate $\rho(ab)$}. From the embeddings, neurons have inputs $\rho(a)$ and $\rho(b)$, and use their non-linearity to calculate $\rho(ab)$. We make this calculation explicit in the 1d case in Appendix~\ref{sec:sign_circuit}. To demonstrate this, we follow the approach taken with embeddings. We define for each representation a \textit{hidden representation subspace} of rank $d^2$ of $\R^{n^2}$, and consider the projection of the hidden layer onto these subspaces.

\textbf{Neurons cluster by representation}. Our neurons cluster into disjoint categories, corresponding to key representations.
This clustering is identical on neuron inputs and outputs. $7$ neurons are `sign neurons': these neurons completely represent $\rho_{sign}(a)$ in the left embedding and $\rho_{sign}(b)$ in the right embedding. On post-activation outputs, they represent some linear combination of $\rho_{sign}(a)$, $\rho_{sign}(b)$, and $\rho_{sign}(ab)$, but not any other representation. $119$ neurons are correspondingly `standard neurons'. The final $2$ neurons are always off.

In Table~\ref{tb:mlp_neurons} we find $88.0\%$ of the variance of standard neurons can be explained by the directions corresponding to $\rho(a), \rho(b)$ and $\rho(ab)$. For sign neurons, this fraction of variance of neurons explained is $99.9\%$. We validate by ablation the residual $12.0\%$ of standard neurons does not affect performance. We hypothesize this term is a side product of the network performing multiplication with a single ReLU, and discuss this multiplication step further in Appendix~\ref{sec:sign_circuit}. Evolution of percentage of MLP activations explained by each representation is presented in Figure~\ref{fig:percent_hidden}.

\begin{table}[h]
\vspace{-0.3cm}
\caption{Percentage of the variance of MLP neurons explained by subspaces corresponding to representations of group elements $a$, $b$ and $ab$. Almost all of the variance of neurons within each key representation cluster is explained by subspaces corresponding to the representation, and all neurons are in a single cluster.}
\vskip 0.15in
\begin{center}
\begin{small}
\begin{sc}
\begin{tabular}{lcccc}
\toprule
Cluster &$\rho(a)$ & $\rho(b)$ & $\rho(ab)$ & Residual\\
\midrule
sign & 33.3\% & 33.3\% & 33.3\% & 0.00\%\\
standard & 39.6\% & 37.1\% & 11.3\% & 12.1\%\\
\bottomrule
\end{tabular}
\end{sc}
\end{small}
\end{center}
\label{tb:mlp_neurons}
\end{table}

\textbf{Only the $\rho(ab)$ component of MLP neurons is important}. The \abbrevalgorithm{} algorithm doesn't make use of $\rho(a)$ or $\rho(b)$ directly to compute $\chi_\rho(abc^{-1})$. We confirm the model too only makes use of $\rho(ab)$ type terms by ablating directions corresponding to $\rho(a)$ and $\rho(b)$ or otherwise in MLP activations and verifying loss doesn't change. 

On the other hand, ablating directions corresponding to $\rho(ab)$ in the key representations severely damages loss. Baseline loss is $2.38 \times 10^{-6}$. Ablating $\rho_{standard}(ab)$ increases loss %by a relative factor of $3.2\times10^6$
to $7.55$, while ablating $\rho_{sign}(ab)$ increases loss %by a factor 380 
to $0.0009$.

\textbf{We may explicitly recover representation matrices from hidden activations}. By changing basis (via Figure~\ref{fig:basis}) on the hidden representation subspace corresponding to $\rho(ab)$, we may recover the matrices $\rho(ab)$. The learned sign representation matrices agree with $\rho_{sign}(ab)$ completely, and the learned standard representation matrices agree with $\rho_{standard}(ab)$ with MSE loss $<10^{-8}$. We cannot recover representation matrices for representations not learned.

%By performing an appropriate change of basis, we can explicitly extract $\rho(ab)$ representation matrices from the hidden layer}. For the key representations, the MLP neurons contain directions corresponding to $\rho(ab), \rho(a)$ and $\rho(b)$. We first project onto the subspace of the hidden layer spanned by only by $\rho(ab)$ directions, and then perform a non-orthogonal change of basis from the hidden layer to representation space. We find the learned standard representation matrices agree with the actual $\rho_{standard}(ab)$ with cosine similarity $99.97\%$, and the learned sign representation matrices agree with $\rho_{sign}(ab)$ completely.

\subsection{Logit Computation}
\label{sec:logit_comp}

\textbf{Maps to the logits are localised by representation}. The unembedding map $W_U$ restricts to each key representation neuron cluster. This restricted map, following a similar approach to Section~\ref{sec:embeddings}, has almost all components in the corresponding output representation subspace. Defining $W_\rho$ as the map from $\rho$-neurons to logits, we find $W_{sign}$ has $99.9\%$ variance explained by output sign representation space, and $W_{standard}$ has $93.4\%$ explained by output standard representation space.

\begin{figure}[!ht]
    \centering
    \includegraphics[width=0.6\columnwidth]{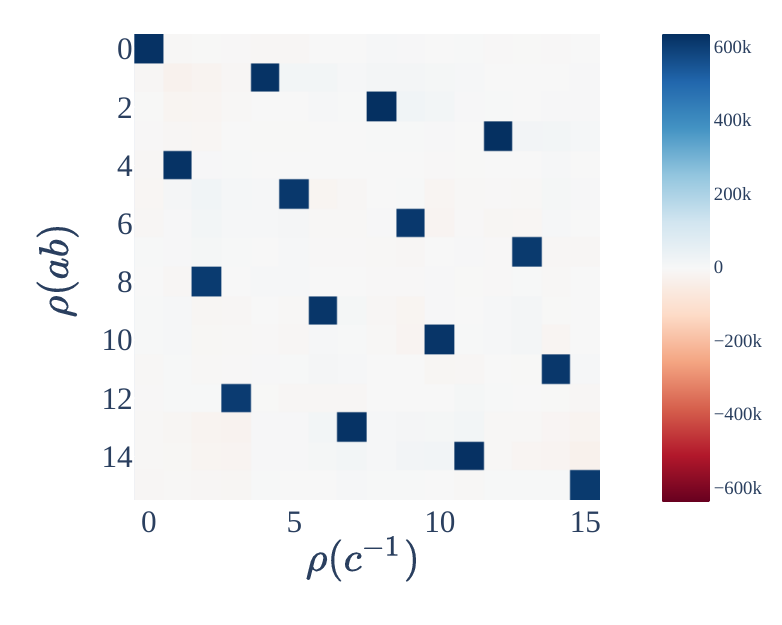}
    \vspace{-0.3cm}
        \caption{The map from the subspace corresponding to $\rho_{standard}(ab)$ in the MLP neurons to logits. We obtain this by changing basis of $W_U$ on both sides, to align with $\rho(ab)$ representation space on the left, and $\rho(c^{-1})$ on the right. This matrix implements step 3 in the \abbrevalgorithm{} algorithm, mapping $\rho(ab)$ to $\chi_\rho(abc^{-1})=tr(\rho(ab)\rho(c^{-1}))$. The sparse and uniform matrix shown corresponds precisely to the trace calculation between two $4\times 4$ matrices as in Equation~\ref{eq:linear_map}.}
    \label{fig:unembedding_matrix}
\end{figure}

\textbf{The linear map in representation basis}. As noted in Section~\ref{algorithm}, the final step of the \abbrevalgorithm{} algorithm may be implemented in a single linear operation (Equation~\ref{eq:linear_map}). Given $\rho(ab)$ is present in MLP neurons, the unembedding need simply learn the inverse representation matrices $\rho(c^{-1})$. We verify the network implements this step as predicted by our algorithm in Figure~\ref{fig:unembedding_matrix}.

%To read off this linear map from weights, we change basis on the input space to be aligned with the $\rho(ab)$ hidden representation space, and change basis on the output space to be aligned with $\rho(c^{-1})$ representation space. The resultant $d^2 \times d^2$ matrix is shown in Figure~\ref{fig:unembedding_matrix} for the standard representation, and encodes precisely the last step of the algorithm.

\subsection{Correctness Checks: Ablations}

In previous sections, we showed various components of the model were well approximated by intermediate terms of the proposed \abbrevalgorithm{} algorithm. To verify these approximations are faithful, we perform two types of additional ablations. We \textbf{exclude} components in the algorithm and verify loss increases, and we \textbf{restrict} to these same components and demonstrate loss remains the same or decreases.

\textbf{MLP neurons.} In Section~\ref{sec:mlp_neurons}, we identified sets of neurons that could be manipulated to recover representation matrix elements $\rho(ab)$. If we replace these neurons with the corresponding representation matrix elements directly, we find loss decreases by $70\%$ (to $7.00\times 10^{-7}$).

\textbf{Unembeddings.} In Section~\ref{sec:logit_comp}, we found $W_U$ is well approximated by $16+1$ directions, corresponding to representation space on the two key representations. If we project MLP neurons to only these directions, ablating the $5.96\%$ residual in $W_U$, we find loss decreases by $12\%$, while if we project to only this residual,  loss increases to $4.80$, random.

\textbf{Logits.} In Section~\ref{sec:logit_attribution} we found observed logits were well approximated by the \abbrevalgorithm{} algorithm in the key representations. We find ablating our algorithm's predictions in the key representations damages loss, to 0.0006 by excluding the sign representation, to 7.23 excluding the standard representation, and to 7.60 excluding both, significantly worse than random. Ablating other directions improves performance.

\subsection{Understanding Training Dynamics using Progress Measures}
\label{sec:progress_measures}

\begin{figure}[!ht]
    \centering
    \begin{subfigure}{}
    \includegraphics[width=0.9\columnwidth]{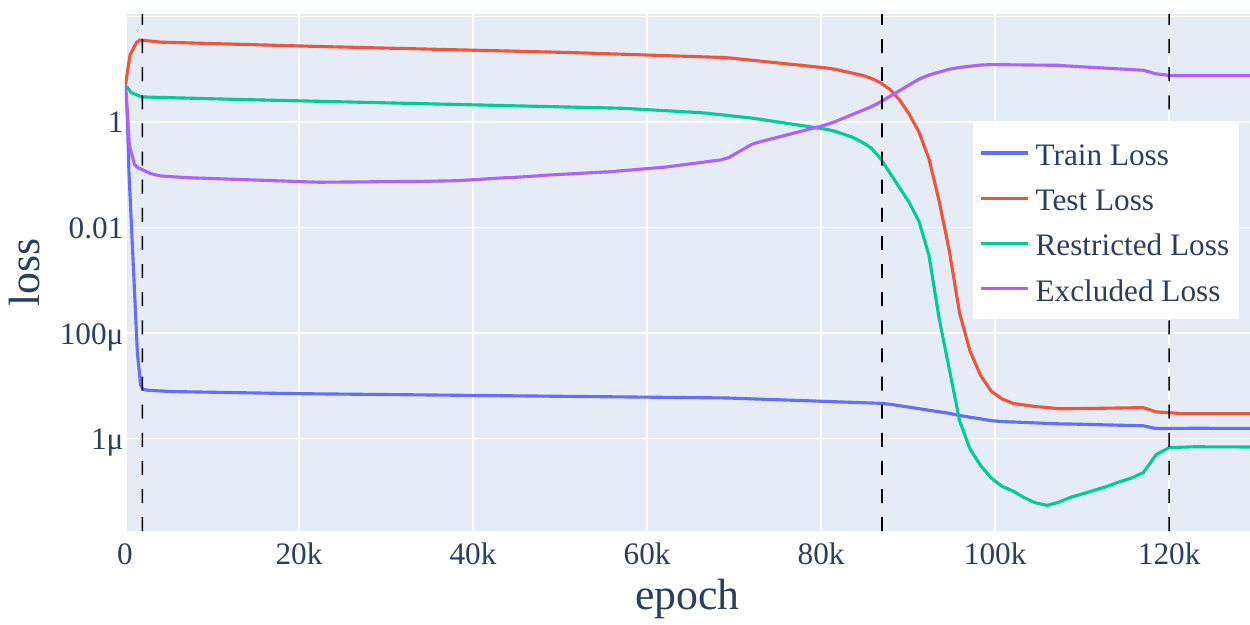}
    \end{subfigure}
    \vspace{-0.5cm}
    \caption{Evolution of the two progress measures over training. The vertical lines delineate 3 phases of training: memorization, circuit formation, and cleanup (and a final stable phase). Excluded loss tracks the progress of the memorization circuit, and accordingly falls during the first phase, rising after during circuit formation and cleanup. Restricted loss tracks the progress of the generalized algorithm, and has started falling by the end of circuit formation. Note that grokking occurs during cleanup, only after restricted loss has started to fall.}
    \label{fig:progress_measures}
\end{figure}

A limitation of prior work on using \textit{hidden progress measures} from mechanistic explanations as a methodology for understanding emergence \cite{nandaProgressMeasuresGrokking2023} is that the technique developed may not generalize beyond one specific task. We demonstrate their results are robust by replicating them in our network trained on $S_5$. 

We argue that the network implements two classes of circuit -- first, `memorizing' circuits, and later, `generalizing' circuits. Both are valid solutions on the training distribution. %Grokking is hard to study, because these are both valid solutions on the training distribution, but look the same. 
To disentangle these, we define two progress measures. \textbf{Restricted loss} tracks only the performance of the generalizing circuit via our algorithm. \textbf{Excluded loss} is the opposite, tracking the performance of only the memorizing circuit, and so is only evaluated on the training data. We find that on our mainline model, training splits into three partially overlapping phases -- memorization, circuit formation, and cleanup. During circuit formation, the network smoothly transitions from memorizing to generalizing. Since test performance requires a general solution and no memorization, grokking occurs during cleanup. Further discussion may be found in Appendix~\ref{app:progress_measures}.

In our mainline experiments, we use weight decay as the primary regularization scheme. Other regularizers are also capable of exhibiting grokking. Our results mirror \cite{nandaProgressMeasuresGrokking2023}: we find models grok generic group composition under dropout, and the methodology of progress measures can too be used to understand grokking in this case.

We sometimes find further phase changes. Figure~\ref{fig:phase_change_grok} demonstrates two \textit{phases of grokking} in a seperate run, caused by learning of different representations at distinct times.

%\subsection{Representation Learning}
%\label{sec:rep_learning}

%Figure~\ref{fig:logit_similarity} shows that the sign representation is learnt early on in training. It is %however unable to solve the task alone. This is because the representation is \textit{non-faithful} (see Definition~\ref{def:faithful_rep}). It however is easy to learn. The standard representation is harder to learn and learnt later in training. It is capable of solving the task alone, though combining it with the sign representation improves loss due to constructive interference on correct answers and destructive interference on incorrect answers. We discuss this observation further in Section~\ref{sec:universality}. These correspond to type 1 and type 2 features according to the taxonomy presented in \cite{daviesUnifyingGrokkingDouble2022}.

%We additionally see in Figure~\ref{fig:logit_similarity}, that while the ``standard\_sign'' representation is not retained in the final model, so not classed as a key representation, it was learned by the network at some point. It appears to be directly traded for the sign representation, perhaps due to the sign representation offering more performance per unit weight?

\section{Universality}
\label{sec:universality}

\begin{table*}[!ht]
\vspace{-0.25cm}
\tiny
\caption{Results from all groups on both MLP and Transformer architectures, averaged over 4 seeds. We find that that features for matrices in the key representations are learned consistently, and explain almost all of the variance of embeddings and unembeddings. We find that terms corresponding to $\rho(ab)$ are consistently present in the MLP neurons, as expected by our algorithm. Excluding and restricting to these terms in the key representations damages performance/does not affect performance respectively.}
    \vspace{0.5cm}
    \centering

\input{data/all_avg.tex}
\label{tb:avg}
\end{table*}

In this section, we investigate to what extent the universality hypothesis \cite{olahZoomIntroductionCircuits2020, liConvergentLearningDifferent2016} holds on our collection of group composition tasks. Here, `features' correspond to irreducible representations of group elements\footnote{Defining a `feature' in a satisfying way is surprisingly hard. \citet{nandaComprehensiveMechanisticInterpretability2022} discusses some of the commonly used definitions.} and `circuits' correspond to precisely how networks manipulate these with their weights.
 
We interpret models of MLP and Transformer architectures (Appendix~\ref{sec:architecture_details}) trained on group composition for seven groups: $C_{113}$, $C_{118}$, $D_{59}$, $D_{61}$, $S_5$, $S_6$ and $A_5$, each on four seeds. We find evidence for \textit{weak universality}: our models are all characterized by a family of circuits corresponding to our \abbrevalgorithm{} algorithm across all group representations. We however find evidence against \textit{strong universality}: our models learn different representations, implying that specific features and circuits will differ across models. 

%that networks will learn precisely the same features and circuits consistently. Instead, we find the specific representations vary, resulting in the specific features and circuits chosen from this family varying. %Even if strong universality fails in general, there is still promise that a `periodic table' of universal features, akin to the representations in our group theoretic task, may exist in general for real tasks. We may draw an analogy to the idea of \textit{equivariance} \cite{olahNaturallyOccurringEquivariance2020}, that features of a similar motif may be learned even if the precise features vary.

%Our tooling permits all the lines of evidence in Section~\ref{sec:reverse_engineer} to be produced. As mechanistic interpretability mostly focuses on post-hoc understanding of trained models, and as analysing training dynamics are mostly not important for the study of universality, we choose to only interpret our final model. Full summary statistics are presented in Appendix~\ref{sec:univ_results}. 

\textbf{All our networks implement the \abbrevalgorithm{} algorithm}. We first argue for weak universality via universality of our algorithm and universality of a \textit{family} of features and circuits involving them. Following the approach of Section~\ref{sec:reverse_engineer}, we understand each layer of our network as steps in the algorithm as presented in Section~\ref{algorithm}. (Table~\ref{tb:avg}). Steps 1 and 3 -- we analyze embedding and unembedding matrices, showing that their fraction of variance explained (FVE) by subspaces corresponding to the key representations is high. Each group has its own family of representations, and each model learns its own set of key representations (i.e. representations with non-zero logit similarity). Where applicable, our metrics track only these key representations of any given model. %$ which vary between models
%non-zero logit similarity. 
For Step 2, we show the MLP activations are well explained by the terms $\rho(a)$, $\rho(b)$, and importantly $\rho(ab)$ in the key representations. Finally, as evidence our algorithm is entirely responsible for performance, we show the final values of the progress measures of restricted and excluded loss. %Results are shown in Table~\ref{tb:avg}. 

\begin{figure}
    \centering
    \begin{subfigure}{}
        \includegraphics[width=0.50\columnwidth]{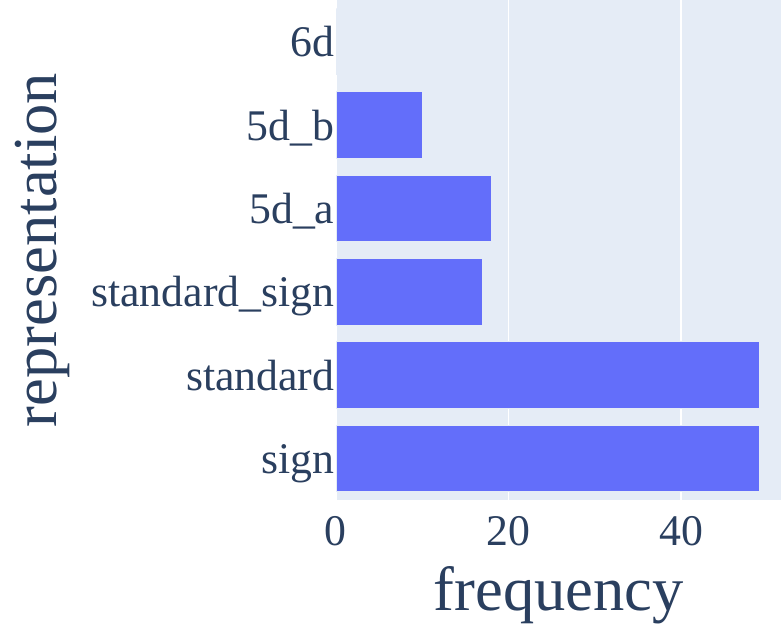}
    \end{subfigure}
    \hfill
    \begin{subfigure}{}
        \includegraphics[width=0.40\columnwidth]{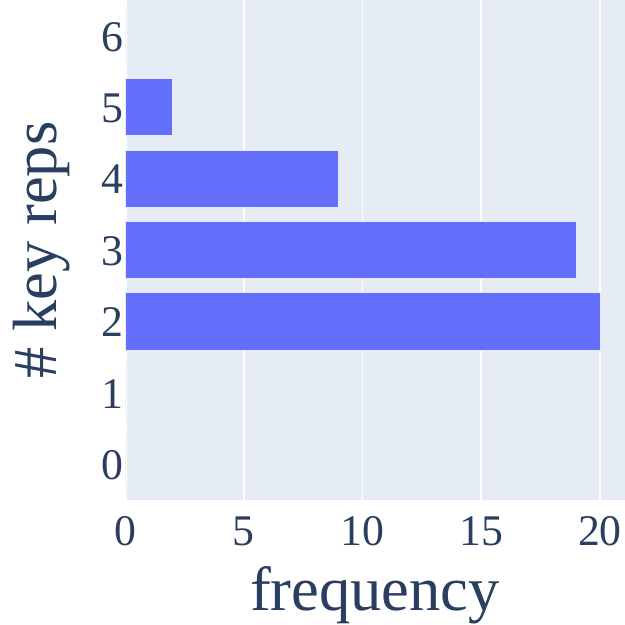}
    \end{subfigure}
    \caption{(\textbf{Left}) The number of times each representation is learned over 50 seeds, for $S_5$ trained on the MLP architecture. We see the 1d sign and 4d standard representations are most commonly learned, standard\_sign (4d), 5d\_a and 5d\_b are learned approximately equally and less often, and 6d is never learned. (\textbf{Right}) The number of key representations of these 50 runs. Most commonly we have two key representations (typically sign and standard), but sometimes we learn more.}
    \label{fig:bars}
    \vspace{-0.4cm}
\end{figure}

\textbf{Specific representations learned vary between random seeds.} Each group has several representations that can be learned. Under strong universality, we would expect the representations learned to be consistent across random seeds when trained on the same group. In general, we do not find this to be true (Figure~\ref{fig:bars}). When there are multiple valid solutions to a problem, the model somewhat arbitrarily chooses between them -- even when the training data and architecture are identical. 

\textbf{It is not the case that networks learn simple representations over complex representations.} If strong universality is true, we hypothesized networks would learn `simple' representations over more complex ones, according to some sensible measure of complexity. 

%We initially replicated the findings of \citet{nandaProgressMeasuresGrokking2023} on cyclic groups of prime order. Representations of such cyclic groups are all 2 dimensional, and entirely analogous, as each representation is a permutation of other representations\footnote{This is a direct consequence of the prime order. See Appendix~\ref{sec:c_reps}.} We therefore take each representation to be of equal complexity, and therefore it is not surprising that across runs an arbitrary set of representations are learned.

\begin{figure}[!t]
    \centering
    \includegraphics[width=0.9\columnwidth]{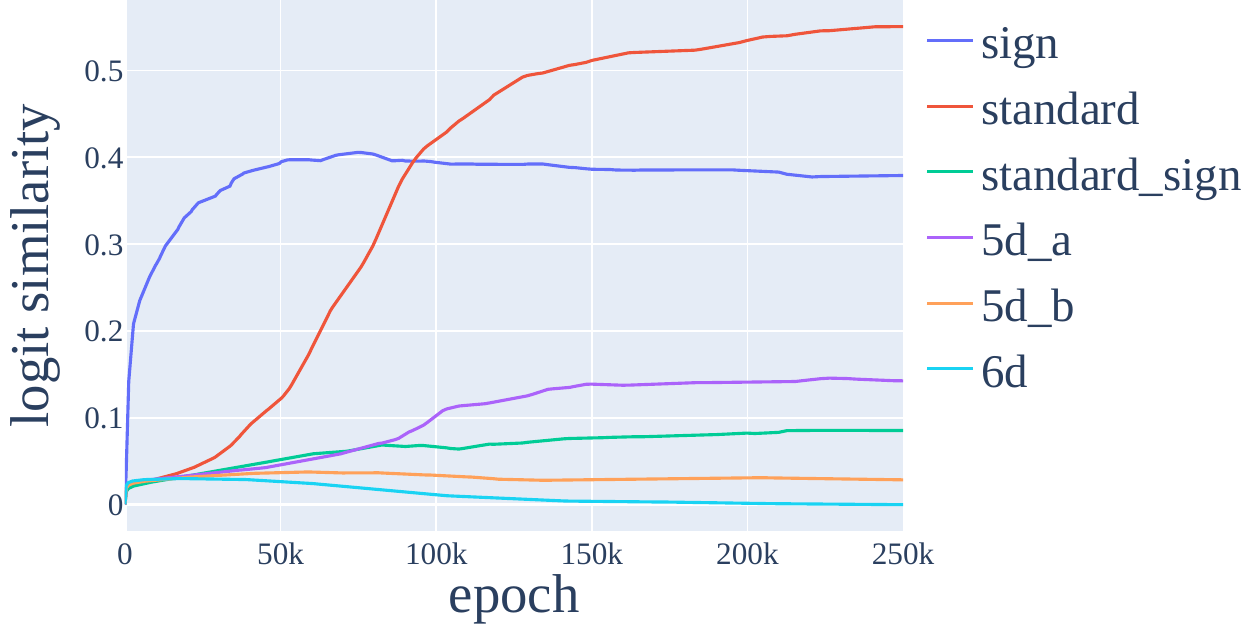}
    \caption{Mean evolution of logit similarity of each non trivial representation of $S_5$ over training averaged over 50 random seeds. We observe the sign representation is consistently learned early in training, and the standard representation is also often learned. Notably, the standard\_sign representation is of comparable complexity to the standard representation, but learned to a lesser degree. }
    \label{fig:s5_rep_learning}
    % \vspace{-0.5cm}
\end{figure}

We naively thought that the complexity of a general representation would correlate with it's dimension\footnote{In particular, we thought a reasonable definition would be the number of linear degrees of freedom in the $n \times d^2$ tensor of flattened representation matrices -- i.e. the rank of the representation subspace of $\R^n$ (from Section \ref{sec:embeddings}).}. For $S_5$, since the 4 dimensional representations are the lowest faithful representations, we expected representations of at most this dimension to be learned, and the model to choose arbitrarily between learning either of the two of them, or both. Empirically, we found this claim to be false. In particular, networks commonly learned higher dimensional representations, as can be seen in Figure~\ref{fig:bars}. We also see in Figures~\ref{fig:bars} and \ref{fig:s5_rep_learning} that the network preferred the standard representation over the standard\_sign representation, when in fact standard\_sign offers \textit{better} performance for fixed weight norm.

While not deterministic, Figure~\ref{fig:bars} shows at least a probabilistic trend between our naive feature complexity and learning frequency, suggesting meaningful measures of feature complexity may exist. One complication here is that, as discussed in Section~\ref{sec:progress_measures}, models are trading off weight against performance. Representations with more degrees of freedom may also offer better performance for fixed total weight norm, so which the model may prefer, and thus which is least complex, is unclear.  

\textbf{Number of representations learned varies.} Across seeds, in addition to different representations being learned, we too find different \textit{numbers} of representations are learned, also shown in Figure~\ref{fig:bars}. This is surprising to us. We additionally find that Transformers consistently learn fewer representations than MLPs, despite having \textit{more} parameters. We view this as further evidence against the strongest forms of circuit and feature universality, and suggests there is a degree of randomness in what solutions models learn. %We take this as mixed evidence for feature  universality. We show that when there are multiple valid solutions to a problem, the model somewhat arbitrarily chooses between them. Even when, naively, the feature complexity is identical and the training data and architecture is identical.

\textbf{Lower dimensional representations are generally (but not always) learned first.} Under any reasonable definition of complexity, the 1d sign representation is simpler than other $S_5$ representations. Figure~\ref{fig:s5_rep_learning} shows that the sign representation is consistently learned first. While it is very easily learned, it also generalizes poorly. In contrast, higher dimensional faithful features are harder to learn but generalize better. These correspond to type 1 and type 3 patterns according to the taxonomy presented in \citet{daviesUnifyingGrokkingDouble2022}. We however \emph{do not find evidence that all representations are learned in strict order of dimension}, against our naive hypothesis's predictions, further evidence against strong universality.

\section{Conclusion and Discussion} 
In this work, we use mechanistic interpretability to show that small neural networks perform group composition via an interpretable, representation theory--based algorithm, across several groups and architectures. We then define progress measures \cite{barakHiddenProgressDeep2023, nandaProgressMeasuresGrokking2023} to study how the internals of networks develop over the course of training. We use this understanding to study the universality hypothesis -- that networks trained on similar tasks learn analogous features and algorithms. We find evidence for weak but not strong forms of universality: while all the networks studied use a variant of the \abbrevalgorithm{} algorithm, different networks (with the same architecture) may learn different sets of representations, and even networks that use the same representations may learn them in different orders. This suggests that reverse engineering particular behaviors in single networks is insufficient for fully understanding that network behavior in general. That being said, even if strong universality fails in general, there is still promise that a `periodic table' of universal features, akin to the representations in our group theoretic task, may exist in general for real tasks. We include further discussion on how this work fits into the wider field of mechanistic interpretability in Appendix~\ref{app:wider_context}. Below, we discuss some areas of future work, with further discussion in Appendix~\ref{app:future_work}.

\textbf{Further investigation of universality in algorithmic tasks.}  We raise many questions in Section~\ref{sec:universality} regarding which representations networks learn. Better understanding the learning rates and generalization properties of features offers a promising direction for future work in understanding network universality. Further understanding the probabilistic nature of which features are learned and at what time may too have future relevance. In particular, lottery tickets \cite{frankleLotteryTicketHypothesis2019} may be present in initialized weights that could allow the learned features of a trained network to be anticipated \textit{before training}. 

\textbf{More realistic tasks and models.} In this work, we studied the behaviour of small models on group composition tasks. However, we did not explore whether our results apply to larger models that perform practical tasks. Future work could, for example, study universality in language models in the style of induction heads in \citet{olssonIncontextLearningInduction2022}.

\textbf{Understanding inductive biases of neural networks.} A key question in the science of deep learning is understanding which classes of algorithms are natural for neural networks to express. Our work suggests that the \abbrevalgorithm{} algorithm is in some sense a `natural' way for networks to perform group composition (Appendix~\ref{sec:inductive_bias}). A more comprehensive understanding of the building blocks of neural networks could speed up interpretability work while helping us better understand larger models. %

\section*{Author Contributions}

\textbf{Bilal Chughtai} was the primary research contributor and lead the project. He wrote the code, ran all experiments, reverse engineered the weights of the network trained on composition on $S_5$ in Section~\ref{sec:reverse_engineer}, and used this to automate the process of reverse engineering many more models in Section~\ref{sec:universality}. He also wrote the paper.

\textbf{Lawrence Chan} provided significant help clarifying, framing and distilling the results, and with editing the final manuscript.

\textbf{Neel Nanda} supervised and mentored the entire project. He developed the complete version of the \abbrevalgorithm{} algorithm based on Sam Marks's original version, and showed that it suffices to use a single faithful representation, and aided in editing the final manuscript.

\section*{Acknowledgments}
We would like to thank Joe Benton and Sam Marks for a conversation at a party that sparked this project and for seeing the connection between representation theory and composition of $S_5$, and additionally to Sam for contributing the core idea of the GCR algorithm.

We are also grateful to Joe Benton, Joseph Bloom, Stephen Casper, Ben Edelman, Jeremy Gillen, Stefan Heimersheim, Adam Jermyn, Cassidy Laidlaw, Eric Michaud and Martin Wattenberg for providing generous and valuable feedback on our manuscript. Over the course of the project, our thinking and exposition was also greatly clarified through correspondence with Spencer Becker-Kahn, Paul Colognese, Alan Cooney and Jacob Merizian.

BC would like to thank the SERI MATS 2.1 program, particularly Joe Collman and Maris Sala, for providing an excellent research environment during the entire project. BC was also supported by SERI MATS for the duration of the project. 

We trained our models using \texttt{PyTorch} \citep{paszkePyTorchImperativeStyle2019} and performed our data analysis using \texttt{NumPy} \citep{harrisArrayProgrammingNumPy2020} and \texttt{Pandas} \citep{mckinneyDataStructuresStatistical2010}. We made use of \texttt{SymPy} \citep{10.7717/peerj-cs.103} to handle permutation group operations, and \texttt{TransformerLens} \citep{nandaTransformerLens2023} to cache internal model activations for interpretability. Our figures were made using \texttt{Plotly} \citep{plotly}.

\bibliographystyle{icml2023}
\bibliography{citations}

\appendix
\onecolumn

\section{Relevance for Mechanistic Interpretality}
\label{app:wider_context}

How might this work influence interpretability work on real models? We view our work as a contribution towards where to direct effort in the field. Mechanistic interpretability focuses on reverse engineering neural networks, and providing mechanistic explanations for model behaviors. 

Recently, the field has been making good progress towards understanding how networks implement behavior in a range of contexts. Initial work successfully reverse engineered neurons in computer vision models, \cite{olahZoomIntroductionCircuits2020, olahFeatureVisualization2017, goh2021multimodal}, finding certain neurons represent interpretable human concepts. Other work has found interpretable components of Transformer language models, such as ‘induction heads’, responsible for copying from earlier in the context window and consequently in context learning \cite{olssonIncontextLearningInduction2022}. \citet{wangInterpretabilityWildCircuit2022} were able to reverse engineer a large subgraph of GPT-2, responsible for successful completions of the indirect object identification task (IOI). \citet{nandaProgressMeasuresGrokking2023} were able to reverse engineer Transformers trained to perform modular addition, and through doing so, understand why these models grokked. Mechanistic interpretability has also been applied to AlphaZero and to a model trained to play Othello \cite{mcgrathAcquisitionChessKnowledge2022, liEmergentWorldRepresentations2023} and has been able to demonstrate these networks too learn human understandable concepts.

Much of this work focuses on a single, small model, sometimes with the explicitly stated goal of generalizing to large foundation models \cite{elhageMathematicalFrameworkTransformer21}. \citet{wangInterpretabilityWildCircuit2022} for instance only investigated one model (GPT-2 small). This is often motivated by the universality hypothesis \cite{olahZoomIntroductionCircuits2020} - that there exist canonical solutions to tasks that networks consistently implement - but investigations into single small models may be too specific. If the universality hypothesis is true, work on small or single models may generalize directly to other/larger models of genuine interest. But if not, the mechanistic interpretability community may be wasting substantial effort and should focus instead on directly interpreting models of genuine interest, or creating tools to automate this process. Better understanding the universality hypothesis is therefore important.

Prior work in mechanistic interpretability has sometimes found similar features and circuits across a range of models.  Different computer vision models were found to contain similar and interpretable “curve detector” and “high low frequency detector” neurons in early layers \cite{olahZoomIntroductionCircuits2020}. Sometimes, the same feature has been found to be computed by different circuits - such as induction heads in Transformer language models, as noted in the appendix here \cite{olssonIncontextLearningInduction2022}. However, no one so far has comprehensively and systematically studied the question of how well mechanistic explanations generalize across models, and how big a weakness focusing on a single model is.

In our work, we sought to answer this question. We chose a toy task, where we were (to our surprise) able to fully enumerate all possible solutions through the different representations which were of varying complexity. Our methods allowed us to inspect which of these ground truth features networks had learned. Through doing so, we found that reverse engineering one model was insufficient to understand behavior in general. Our mainline S5 model only gave us insights into two of the possible circuits used to solve the task (corresponding to the sign and standard representations), out of a possible six. Only after studying many more models were we able to observe all the different mechanisms used to implement the single behavior.

We view our work as a proof of concept that by reverse-engineering circuits in many models, one can build a comprehensive periodic table of features that permits understanding of how networks implement behavior in general. Practically speaking, we then suggest that those studying model behaviors should perform “robustness checks” in many models to truly understand all possible mechanisms behind behavior. This may have future relevance to auditing models via mechanistic interpretability. There exist resources that permit the study of universality in language models already, such as MultiBert \cite{sellamMultiBERTsBERTReproductions2022}, which offers a set of similar models trained on different random seeds, much like our models. One could begin by studying the IOI circuit \cite{wangInterpretabilityWildCircuit2022} in these models, and examining whether the same mechanism is universally learned, and if not, how large the family of possible mechanisms truly are.	

\section{Similarities and differences with prior work on reverse engineering modular addition}

Here, we summarize the prior work of \citet{nandaProgressMeasuresGrokking2023} that we build on, and detail where our experimental approaches differ. We note our contributions differ in that we use our mechanistic understanding to study universality. The authors train a one-layer Transformer model, of same type as we use (Section~\ref{sec:architecture_details}) on modular addition. They find strong evidence it performs a completely understandable algorithm involving discrete Fourier transforms of the two inputs at various frequencies, and then makes use of various trigonometric identities to combine these. The key result which we generalize is that given inputs $a$ and $b$ the network computes $\cosp{\omega(a+b-c)}$ for each possible output $z$ over some fixed set of frequencies $\omega$. Taking the argmax of this expression over $c$ gives the correct answer. One can track the progress of this computation faithfully through the Transformers activations and weights.

Using this mechanistic understanding, the authors define the concept of a `progress measure' that underlies the emergent behavior of grokking, a qualitative and discontinuous change in model behavior. They find that the training history of the model can be separated into three stages. First, the model memorizes the training data. Then, the circuit components for the general algorithm form smoothly. Finally, the memorized algorithm is cleaned up and removed as it is more complex and not favored by weight decay. Grokking occurs during cleanup, at the critical point after which the learned general algorithm is competitive with the `memorized' algorithm -- performance of the general algorithm is heavily hampered by `noise' from the memorized algorithm. Crucially, the progress measures show that the components responsible for grokking arise before the sharp discontinuity in test loss.

We follow this approach closely. Our techniques in Section~\ref{sec:reverse_engineer} are heavily inspired by Nanda et al.'s approach. Our precise analysis though differs substantially. Fourier transforms are elegant, but specific to the modular addition task. We instead work with representation matrices, and subspaces.

On modular addition of 113 elements, i.e. group composition on $C_{113}$, we are able to replicate their results in our framing. As discussed in Section~\ref{algorithm}, their algorithm maps precisely onto our \abbrevalgorithm{} algorithm, and both approaches may be used to understand the cyclic group task. The mapping of their findings onto ours is fairly clear for embeddings, unembeddings and logits. For MLP neurons, they found that most neurons were well explained by a quadratic form of sinusoidal functions of the 9 terms within a single frequency. This quadratic form shared coefficients in such a way such that this had 2 redundant degrees of freedom, giving 7 terms. In our case, MLP neurons contain information pertaining to $\rho(a)$, $\rho(b)$ and $\rho(ab)$. In the special case of cyclic representations (see Appendix~\ref{sec:c_reps}), each of these terms has 2 degrees of freedom by antisymmetry. Adding a constant gives precisely the same seven terms.

\section{Architecture Details}
\label{sec:architecture_details}

Our mainline model is trained on $40\%$ of all $n^2$ entries in the multiplication table of the group. We use full batch gradient descent. We use weight decay with $\lambda = 1$, and the AdamW optimizer, with learning rate $\gamma = 0.001, \beta_1=0.9$ and $\beta_2=0.98$. We perform $250,000$ epochs of training. As there are only $n^2$ possible input pairs, we evaluate test loss and accuracy on all pairs of inputs not used for training. 

\subsection{MLP}
\label{app:mlp}

Our MLP architecture is summarized in Figure~\ref{fig:architecture}. Inputs $a$ and $b$ are encoded as $n$ dimensional one-hot vectors. Each one-hot vector is embedded with $d = 256$. These are concatenated to form a $512$ dimensional vector, which is fed into a $h=128$ linear layer, with no bias term. \footnote{Emperically, we found adding a bias made little difference to our results, though we hypothesize that training models with a bias may improve the model's ability to perform matrix multiplication of activations, and hence interpretability.} The output is mapped via an unembedding linear map, $W_U$, to $n$ logits, corresponding to each of the $n$ group elements. We did not tie the left embedding, right embedding or unembedding matrices. This is a simplified version of the Transformer architecture used by \citet{nandaProgressMeasuresGrokking2023} (described below) which removes attention. Attention is both empirically irrelevant in this prior work, and not predicted to be necessary by our algorithm. The form of logits is therefore

\begin{center}
\begin{BVerbatim}
Logits = W_U @ ReLU( W_MLP @ [W_left @ a, W_right @ b])
\end{BVerbatim}
\end{center}

Note that the embedding matrices and linear layer have no non-linearity between them. When interpreting model calculations we will tie these matrices, and think of the $a$ and $b$ embeddings as the result of passing inputs through both layers. This methodology is inspired by \cite{elhageMathematicalFrameworkTransformer21}, \footnote{Here, the authors tie Transformer's Q and K matrices, and O and V matrices for the same reason.}. The remainder of the operation of the linear layer is then to add these two `total' embeddings and pass them through a ReLU. That is, 
\begin{center}
\begin{BVerbatim}
Logits = W_U @ ReLU( W_a @ a + W_b @ b)
\end{BVerbatim}
\end{center}
where
\begin{center}
\begin{BVerbatim}
W_a = W_MLP[:d, :] @ W_left       W_b = W_MLP[d:, :] @ W_right
\end{BVerbatim}
\end{center}

\subsubsection{Choice of network size}

We note this architecture is over parameterized for our tasks. Smaller networks, with fewer parameters, often struggled to generalize consistently due to optimization issues. We chose a hidden layer size of 128 to avoid these. We do not think the choice of network size generally affected our results. To verify this, we repeated our mainline $S_5$ experiment many more times, on networks with hidden size ranging from 32 to 256. Of those that did generalize, we saw the \abbrevalgorithm{} algorithm was consistently implemented. We did not see a noticeable effect of network parameter count on which representations were learned. Interestingly, networks consistently learned the sign representation early on, even if they did not successfully generalize later. Sometimes, a generalized network with a small hidden layer would throw away the sign representation late in training to make room for another, higher dimensional, representation, with more generalization power. 

\subsection{Transformer}

Our Transformer architecture for other runs is a decoder only architecture is based on \citet{vaswaniAttentionAllYou2017}. It is identical to the set up for mainline experiments in \citet{nandaProgressMeasuresGrokking2023}. The input to the model is of the form ``a b ='',
where a and b are encoded as $n$-dimensional one-hot vectors, and `=' is a special token above which
we read the output c. We use a one-layer ReLU Transformer, token embeddings with d = 128, learned positional embeddings, 4 attention heads of
dimension $d/4 = 32$, and $n = 512$ hidden units in the MLP. At points we analyze it's embedding $W_E$, MLP layer, and map to logits $W_L = W_U W_{out}$, ignoring the residual skip connection, which we find empirically is not utilized significantly for our tasks.

\section{Mathematical Representation Theory}
\label{sec:appendix_rep_theory}

In this section we present the results from group, representation, and character theory we make use of. We begin by motivating our use of representation theory in this context. Groups are an abstraction of the idea of symmetry. In practice though, groups are not purely abstract objects, and tend to arise due to their action on other things. Often, these things are naturally attached to some vector space $V$, such that $G$ gives rise to a linear action $\rho$ on $V$, which we call a representation.

Representation theory appears in several physical systems and is of fundamental importance to science. While groups encode the symmetries of physical systems, representations prescribe the set of possible actions of these symmetries on physical vector spaces. For instance, the representation theory of the particular Lie groups encoding symmetry transformations of spacetime determine the particles predicted by the standard model, which we observe in the universe.

\begin{definition}
A linear representation $\rho$ is a group homomorphism $\rho: G \to GL(V)$ where $GL(V)$ denotes the general linear group of some vector space $V$ over a field $\mathbb{F}$, the set of linear maps on $V$. 
\end{definition}

We focus on real representations, i.e. group homomorphisms $\rho: G \to GL(\R^d)$, the set of real invertible $d \times d$ matrices.  We give some concrete examples of such representations of particular groups in Section~\ref{sec:explicit_reps}. We hypothesize representations are a natural way for a neural networks to implement operations on group elements. Representing group elements in a linear algebra theoretic manner seems like it would be advantageous to a networks natural operations of matrix multiplication and addition. We discuss this observation further in Appendix~\ref{sec:inductive_bias}.

\begin{definition}
\label{def:irrep}
Let $\rho: G \rightarrow GL(V)$ be a linear representation. $\rho$ is said to be \textbf{irreducible} if $\rho$ has no $G$-stable subspace. That is, there is no subspace of $V$ on which $\rho$ defines a sub-representation of $G$.
\end{definition}

From now on, we will use the term \emph{irrep} to refer to irreducible representations. Irreps are the key object of interest. This is due to Maschke's Theorem.

\begin{theorem}
\label{th:maschke}
\textbf{(Maschke)} Every representation of a finite group G is a direct sum of irreducible representations. That is, there exists some basis in which all representation matrices are block diagonal, where the block sizes $d_1, \dots, d_k$ are the same for all $\rho(g)$ with $g\in G$.
\end{theorem}

\begin{example}
Every group $G$ has a representation for any dimension d mapping each group element to identity matrix $I_d$. This is the direct sum of $d$ one-dimensional irreducible representations named the `trivial' representation, given by $\rho(g) = 1$ for all $g\in G$.
\end{example}

This representation isn't practically useful, as the network can not use these representations to perform calculations on group elements. We will often exclude the trivial representation and refer to \emph{non-trivial representations}. There are a finite number of these due to

\begin{theorem}
\label{th:sum_of_dims}
Let $G$ be a group of order $n$ and let $\rho_i$ be distinct (up to isomorphism) irreducible representations of G over some splitting field $\mathbb{F}$. Let $d_i$ be the dimension of $\rho_i$, and $r$ be the number of irreducible representations. Then $n = d_1^2 + \dots + d_r^2$.
\end{theorem}
%[todo: define splitting field https://groupprops.subwiki.org/wiki/Splitting_field]

Some representations are more useful to the network than others:
\begin{definition}
\label{def:faithful_rep}
A representation $\rho$ is said to be \textbf{faithful} if different elements $g$ of $G$ are represented by distinct linear maps $\rho(g)$. In other words, the group homomorphism $\rho: G\rightarrow GL(V)$ is injective.
\end{definition}

Faithful representations are the most useful to the network, though we will often see networks also make use of lower degree non-faithful representations too. 

Character theory forms an important part of representation theory, and will be important to our use case. 
\begin{definition}
\label{def:character}
Let V be a finite-dimensional vector space over a field $\mathbb{F}$ and let $\rho : G \to GL(V)$ be a representation of a group G on V. The \textbf{character} of $\rho$ is the function $\chi_\rho : G \to \mathbb{F}$ given by $\chi_\rho (g) = \tr \rho(g)$, the trace of the representation matrix.
\end{definition}

We now present some useful facts about characters. Character's are \textit{class functions} -- that is, they take a constant value on each conjugacy class of the group. Note too that 

$$ \chi(g^{-1}) = \overline{\chi(g)}$$

In the case of real representations this implies

$$ \chi(abc^{-1}) = \chi((abc^{-1})^{-1}) = \chi(c(ab)^{-1}) = \chi((ab)^{-1}c) $$

where in the final step we used the cyclic property of trace. $\chi((ab)^{-1}c)$ is naively an alternative valid computation the network could use to compute correct answers, and this shows it is equivalent to the \abbrevalgorithm{} algorithm.

\begin{theorem}
\label{thm:trace}
Let $G$ be a group, and $\rho: G \rightarrow GL(\R^d)$ a real representation of it of dimension $d$. For $g \in G$, $\chi_\rho(g) = \tr{\rho(g)}\leq d$ with equality iff $\rho(g)=I$.
\begin{proof}
Let $|G| = n$. Since $\rho$ is a group representation, and the order of elements in a group divide $n$, $\rho(g)^{n}=I$ for all $g$. The eigenvalues of $\rho(g)$ are therefore $n$'th roots of unity, so each character is a sum of roots of unity. By the triangle inequality, the claim holds.
\end{proof}
\end{theorem}

\begin{theorem}\textbf{(Schur's Orthogonality Relation of Characters)}
\label{th:schur}
The space of complex-valued class functions of a finite group G is endowed with a natural inner product, given by

$$ \langle \alpha, \beta \rangle = \frac{1}{|G|}\sum_{g\in G}\alpha(g)\overline{\beta(g)} $$

where $\overline{\beta(g)}$ denotes the complex conjugate. With respect to this inner product, the irreducible characters form an orthonormal basis for the space of class functions, yielding the orthogonality relation

$$ \langle \chi_i, \chi_j \rangle = \begin{cases}
      0 & \text{if $i\neq j$}\\
      1 & \text{if $i=j$} 
    \end{cases}       $$
\end{theorem}

\begin{theorem}\textbf{(Schur's Orthogonality Relation of Matrix Elements)}
\label{th:schur_matrix}
Let $\rho_\lambda$ be irreducible representations of a finite group $G$ of dimension $d_\lambda$ with matrix presentations $\Gamma^\lambda_{ij}$. Without loss of generality, we may assume $\Gamma^\lambda$ is unitary, as any matrix representation is equivalent to a unitary representation. 

Then 
$$ \sum_{g \in G} \overline{\Gamma^\lambda(g)}_{ij}\Gamma^\mu_{i'j'} =  \delta^{\lambda_\mu}\delta_{ii'}\delta_{jj'} \frac{|G|}{d_\lambda} $$

\end{theorem}
Note that the overbar denotes a complex conjugate, and the unitarity assumption only affects the constant, not the orthogonality.

\subsection{Explicit Groups and Representations}
\label{sec:explicit_reps}

Our methods for reverse engineering networks require mechanistic understanding of the precise form of representations. Here, we describe the irreducible representation matrices for particular groups. The classification of irreducible representations for any given group requires some machinery not presented here, and which we don't require for the purposes of our work. We just state the key results.

\subsubsection{Irreducible Representations of the Cyclic Group}
\label{sec:c_reps}

The cyclic group $C_n$ encodes rotational symmetries of an n-gon. Over the reals, the irreducible representations of $C_n$ fall into three classes. Note that Theorem~\ref{th:sum_of_dims} does not apply here as $\mathbb{C}$ is a splitting field for $C_n$, but $\R$ is not.

% https://ncatlab.org/nlab/show/cyclic+group
% https://www.uni-math.gwdg.de/tammo/rep.pdf
\begin{enumerate}
    \item the 1-dimensional trivial representation $\mathbf{1}$
    \item the 1-dimensional sign representation $\mathbf{1_{sgn}}$, which only appears if the group order is even.
    \item the 2-dimensional standard representations $\mathbf{2_k}$ of rotations in the Euclidean plane by angles that are integer multiples of $\frac{2\pi k}{n}$ for $k\in \mathbb{N}$ $0 < k < n/2$. The representation matrices may be written explicitly as 

$$ 
\rho_k(x) = 
\begin{pmatrix}
\cosp{\frac{2\pi k}{n}x} & -\sinp{\frac{2\pi k}{n}x} \\
\sinp{\frac{2\pi k}{n}x} & \cosp{\frac{2\pi k}{n}x} 
\end{pmatrix}  
$$

\end{enumerate}

Note the complex representations are much simpler, consisting of the $n$'th roots of unity. The sign representation appears then due to $-1$ being a root of unity iff $n$ even.  For $k=n/2$, the 2d representation is the direct sum of two copies of the sign representation, so is not irreducible, and for $k>n/2$ we have the isomorphism $\mathbf{2_{n-k} \backsimeq \mathbf{2_k}}$. 

\subsubsection{Irreducible Representations of the Dihedral Group}
\label{sec:d_reps}

We focus on dihedral groups $D_{n} = \langle r,s | r^n = s^2 = e, srs = r^{-1} \rangle$, with $n$ odd. These encode all symmetries of an n-gon, rotational and reflectional. The representations of these groups are much the same as those of cyclic groups, and fall into three categories.

\begin{enumerate}
    \item the 1-dimensional trivial representation $\mathbf{1}$
    \item the 1-dimensional sign representation $\mathbf{1_{sgn}}$, mapping $\langle r \rangle$, i.e. rotations, to $1$, and the coset, i.e. reflections to $-1$.
    \item the 2-dimensional standard representations $\mathbf{2_k}$, corresponding to rotations and reflections in the Euclidean plane.

$$ 
\rho_k(r^l) = 
\begin{pmatrix}
\cosp{\frac{2\pi k}{n}l} & -\sinp{\frac{2\pi k}{n}l} \\
\sinp{\frac{2\pi k}{n}l} & \cosp{\frac{2\pi k}{n}l} 
\end{pmatrix}  
$$

$$ 
\rho_k(r^ls) = 
\begin{pmatrix}
\cosp{\frac{2\pi k}{n}l} & \sinp{\frac{2\pi k}{n}l} \\
\sinp{\frac{2\pi k}{n}l} & -\cosp{\frac{2\pi k}{n}l} 
\end{pmatrix}  
$$

\end{enumerate}

\subsubsection{Irreducible Representations of the Symmetric Group}
\label{sec:s_reps}

Our mainline experiments involve the permutation, or symmetric, group of 5 elements, denoted $S_5$. We denote general permutation groups of n elements $S_n$. This is an interested group to look at due to Cayley's Theorem:

\begin{theorem}
\label{th:cayley}
\textbf{(Cayley)} Every group is isomorphic to a subgroup of a symmetric group. 
\end{theorem}

% https://en.wikipedia.org/wiki/Representation_theory_of_the_symmetric_group

\begin{table}[!ht]
\caption{The lowest degree irreps for $S_n$ for $n\geq 7$, and their dimension. For $n \leq 7$, additional symmetries give rise to other low dimensional irreps on top of these.}
\label{tb:s_reps}
\vskip 0.15in
\begin{center}
\begin{small}
\begin{sc}
\begin{tabular}{lccr}
\toprule
\textbf{$S_n$ irrep} & \textbf{Dimension} \\
\midrule
Trivial & $1$   \\                                        
Sign & $1$ \\
Standard  & $n-1$  \\ 
Standard $\otimes$ Sign & $n-1$  \\                      
\bottomrule
\end{tabular}
\end{sc}
\end{small}
\end{center}
\vskip -0.1in
\end{table}

We list the lowest dimensional irreps of $S_n$ in Table~\ref{tb:s_reps}. These may be fairly easily constructed. We constructed trivial irreps in Appendix~\ref{sec:appendix_rep_theory}, but to recap, this just maps every group element to the scalar $1$.

The sign representation are a set of $1\times 1$ matrices representing a kind of parity. Permutations may be decomposed as a (non unique) sequence of swaps. The parity of this number of swaps is in fact well defined, and defines a subgroup of the symmetric group named the alternating group. Mapping this alternating group to $+1$, and the coset to $-1$ gives the sign representation. In general, any group containing a subgroup of index 2 is naturally endowed with a sign representation in a similar manner.

Next is the standard representation. This is essentially the set of permutation matrices -- $n \times n $ square binary matrices, with only one 1 in each row and column, and 0s elsewhere. This representation has dimension $n$, though, not $n-1$. This is because it turns out to be reducible. Recalling Definition~\ref{def:irrep}, this has an invariant subspace under the action of $G$, spanned by the vector sum of all basis elements. The irreducible representations recovered are the standard and trivial representations.

Standard $\otimes$ Sign denotes the tensor product of the standard and sign representations, which is just their matrix product as the sign representation is 1 dimensional.

$S_5$ has three higher degree representations, which I denote 5d\_a, 5d\_b, 6d. We omit their construction here.

\subsubsection{Irreducible Representations of $A_5$}
\label{sec:a_reps}

As a subgroup of $S_5$, $A_5$ inherits representations from $S_5$. However, the six dimensional representation of $S_5$ becomes reducible, splitting into two three dimensional irreps of $A_5$. We omit details here.

\section{Additional Reverse Engineering of Mainline Model}
\label{app:reverse_engineer}

Here we give further evidence our mainline model trained on $S_5$ performs the \abbrevalgorithm{} algorithm as detailed in Section~\ref{algorithm}, and give further details regarding our methods.

\begin{figure}[!ht]
     \centering
     \begin{subfigure}{}
         \includegraphics[width=0.49\textwidth]{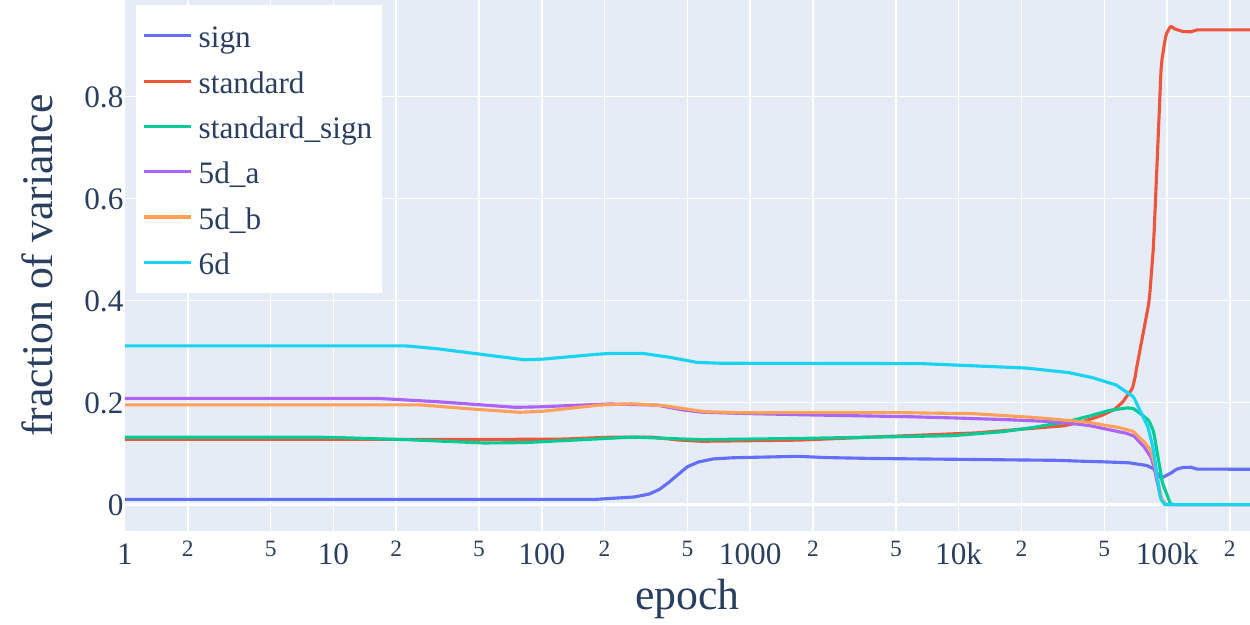}
     \end{subfigure}
     \begin{subfigure}{}
         \centering
         \includegraphics[width=0.49\textwidth]{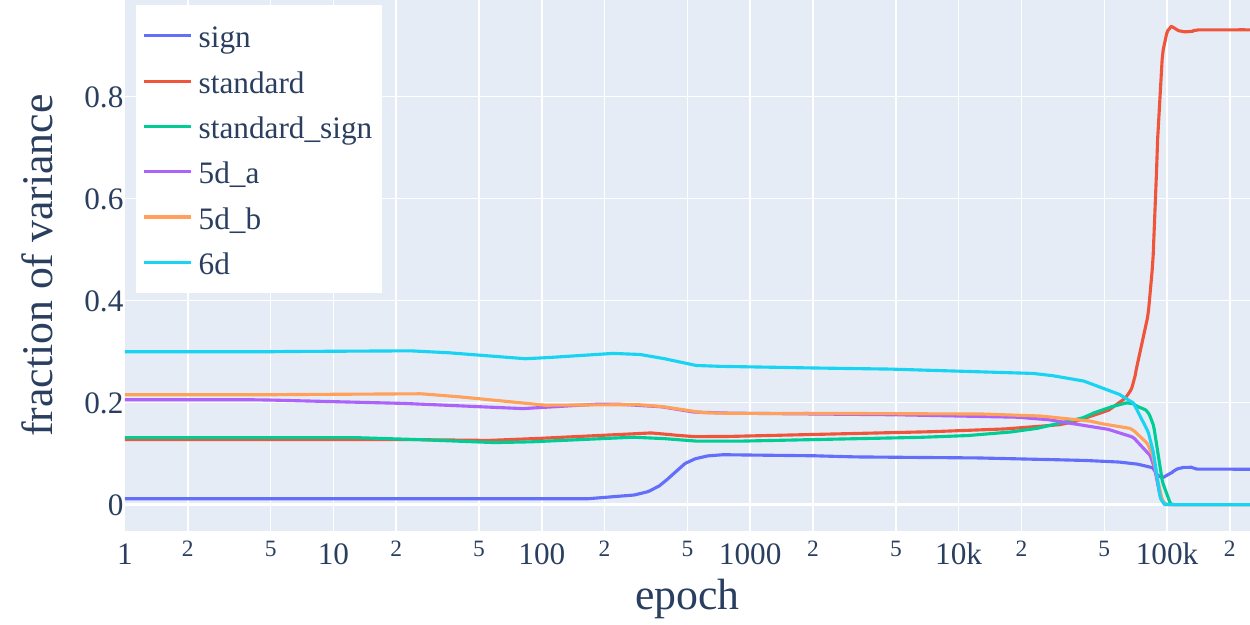}
     \end{subfigure}
     \begin{subfigure}{}
         \includegraphics[width=0.49\textwidth]{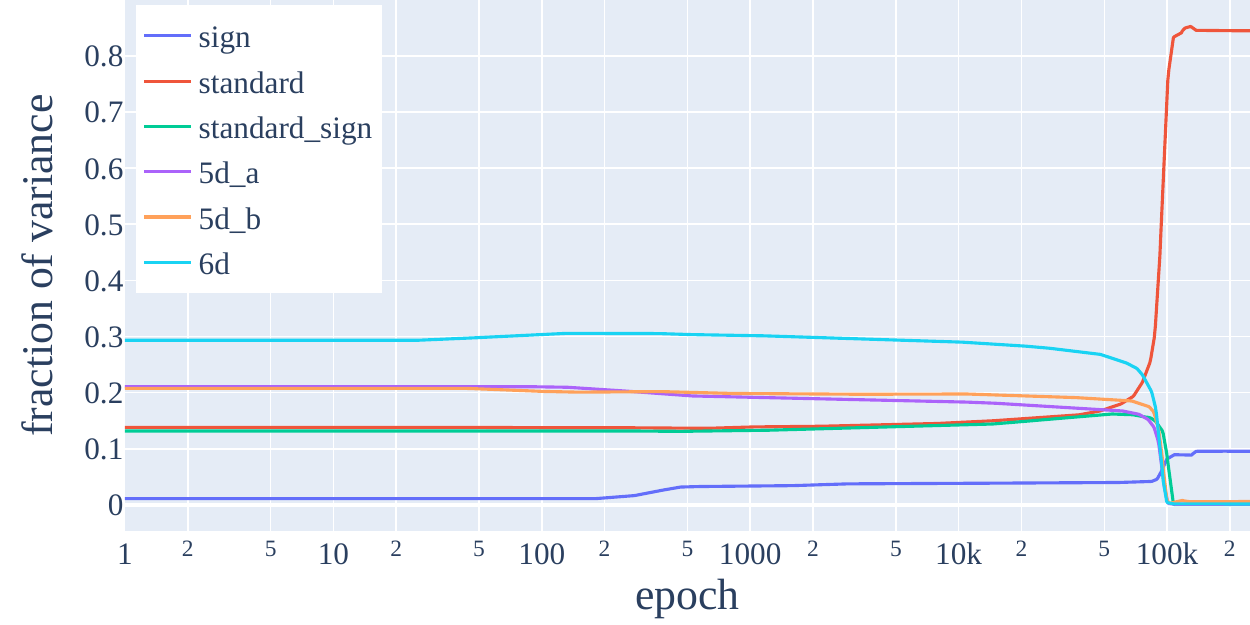}
     \end{subfigure}
    \caption{Evolution of the fraction of the left embedding (\textbf{top left}), right embedding (\textbf{top right}), and unembedding (\textbf{bottom}) explained by $\rho(a), \rho(b)$ and $\rho(c^{-1})$ respectively. Representations are learned suddenly and at approximately the same time across all the embeddings, evidence that they are learned as part of the \abbrevalgorithm{} algorithm. As the representation spaces form an orthogonal decomposition of $\R^n$, the terms will always add up to $1$, so we draw the reader's attention to the sparsity over embeddings. At initialization, each representation explains $d^2/|G|$ of the embedding due to randomness.}
    \label{fig:percent_embeds}
\end{figure}

\begin{figure}[!ht]
     \centering
     \begin{subfigure}{}
         \includegraphics[width=0.49\textwidth]{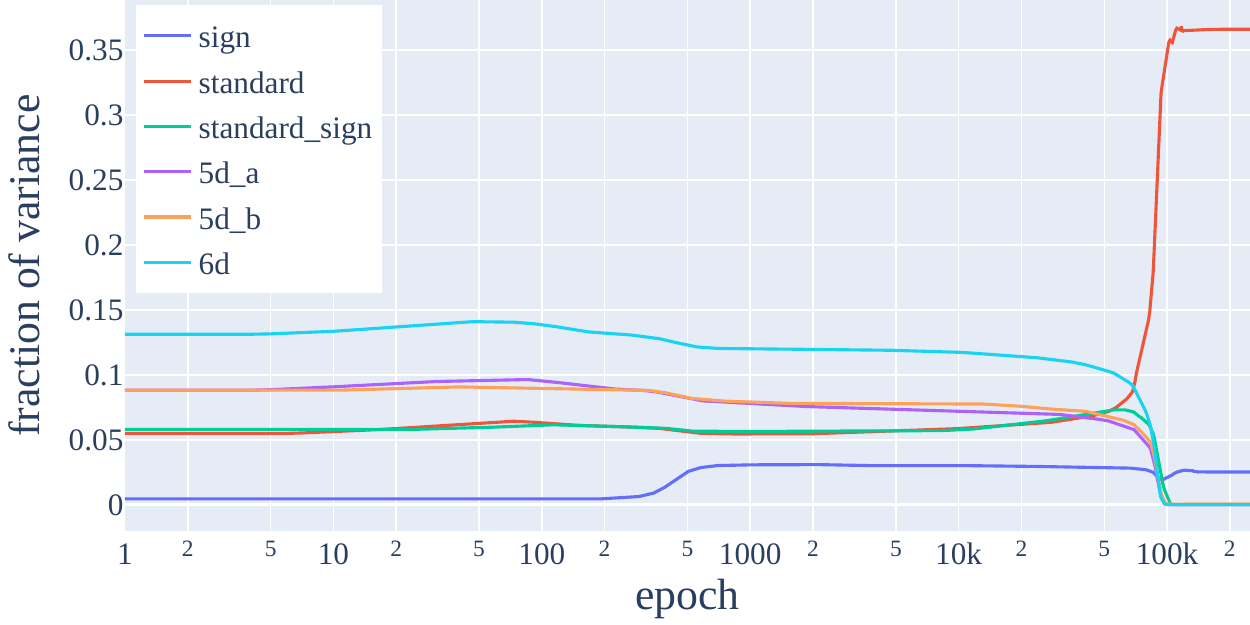}
     \end{subfigure}
     \begin{subfigure}{}
         \centering
         \includegraphics[width=0.49\textwidth]{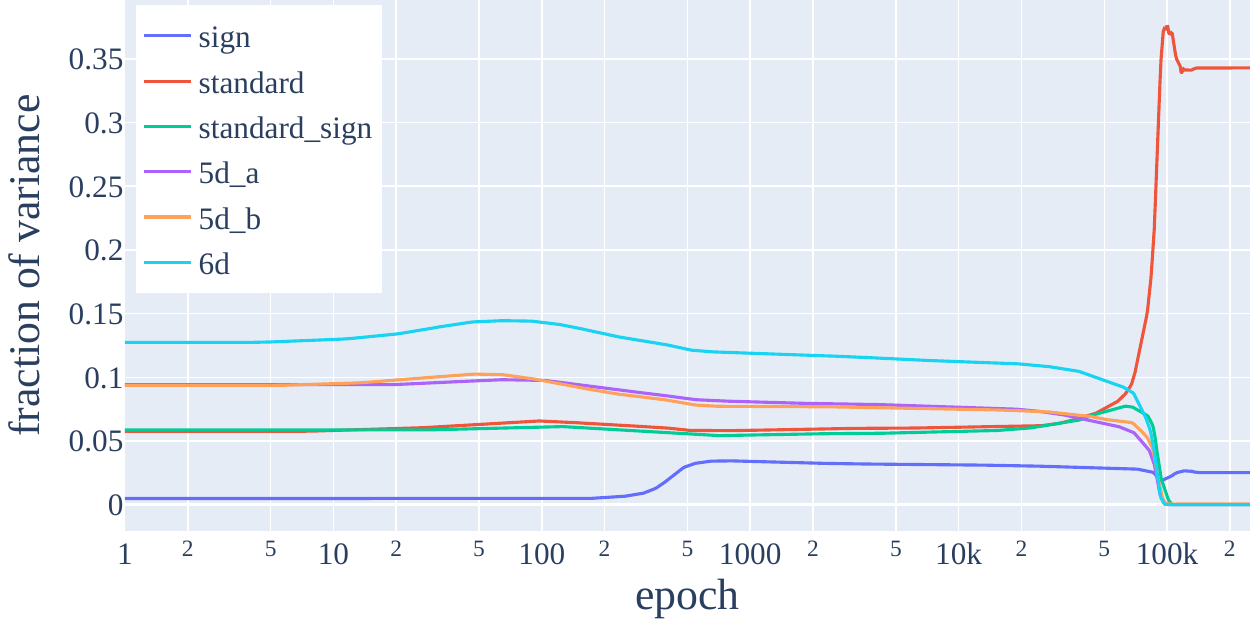}
     \end{subfigure}
     \begin{subfigure}{}
         \includegraphics[width=0.49\textwidth]{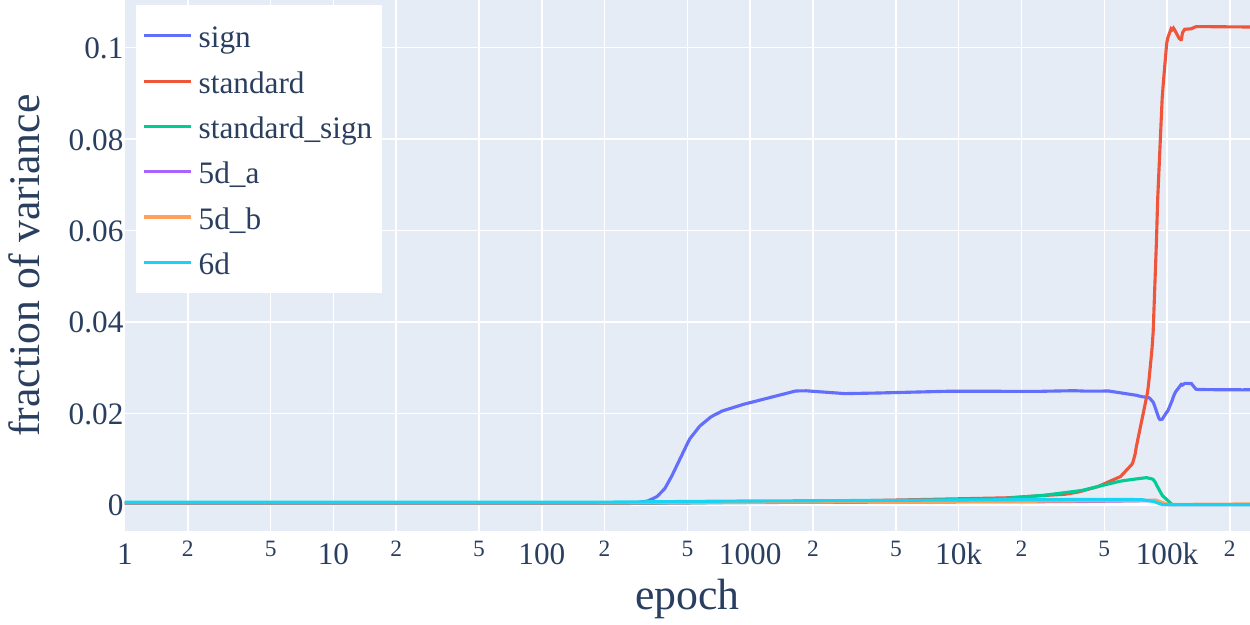}
     \end{subfigure}
     \begin{subfigure}{}
         \includegraphics[width=0.49\textwidth]{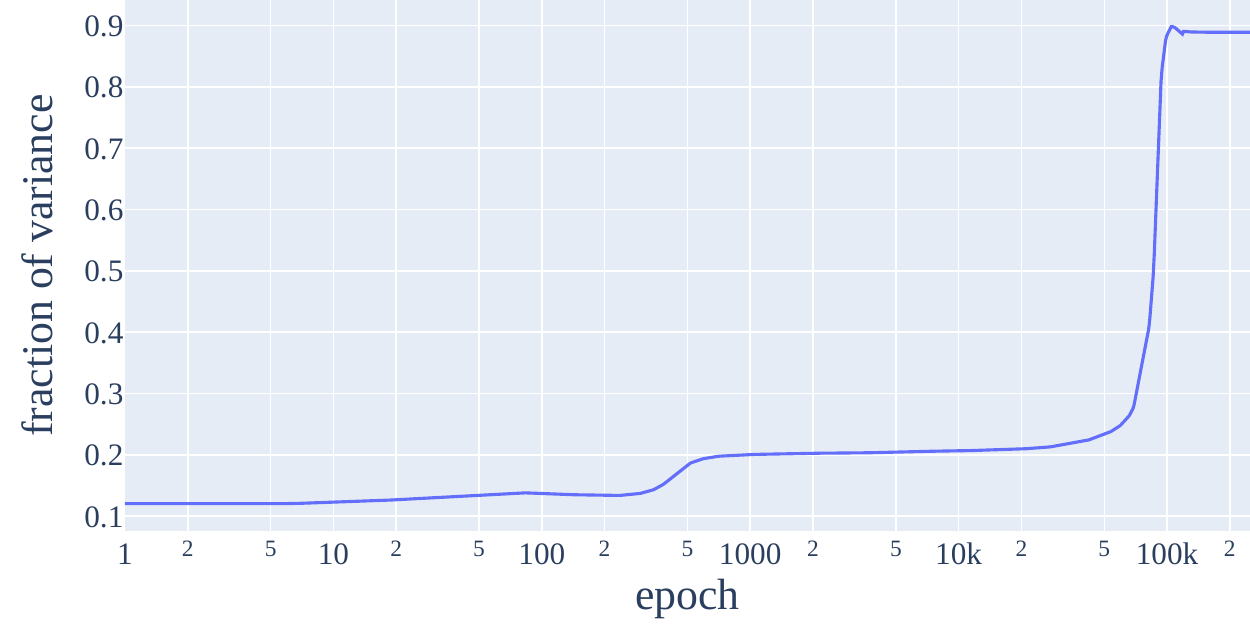}
     \end{subfigure}
    \caption{Evolution of the fraction of the MLP neurons explained by $\rho(a)$ (\textbf{top left}), $\rho(b)$ (\textbf{top right}), $\rho(ab)$ (\textbf{bottom left}), and the sum of all three over all representations (\textbf{bottom right}). These track the same timing as representation learning in the embeddings and unembeddings, further evidence for our algorithm. Note that in order to perform step 2 in the \abbrevalgorithm{} algorithm, $\rho(ab)$ must be calculated. If a representation has $\rho(a)$ and $\rho(b)$ represented but \textit{not} $\rho(ab)$ then the representation has not been learned.}
    \label{fig:percent_hidden}
\end{figure}

\begin{figure}[!ht]
\begin{center}
\centerline{\includegraphics[width=\columnwidth]{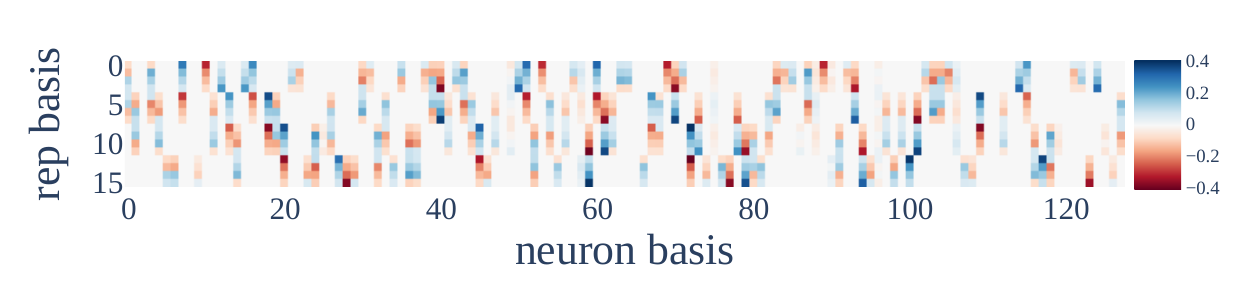}}
\caption{Change of basis matrix from projected MLP space standard representation space. Note some neurons correspond to blocks of 4 cells in the representation basis -- these correspond to standard representation matrix rows. Neurons in other clusters can be explicitly seen as being off in this change of basis matrix.}
\label{fig:basis}
\end{center}
\end{figure}

\newpage

\subsection{Progress Measures}
\label{app:progress_measures}

\begin{figure}[!ht]
    \begin{subfigure}{}
    \includegraphics[width=0.49\textwidth]{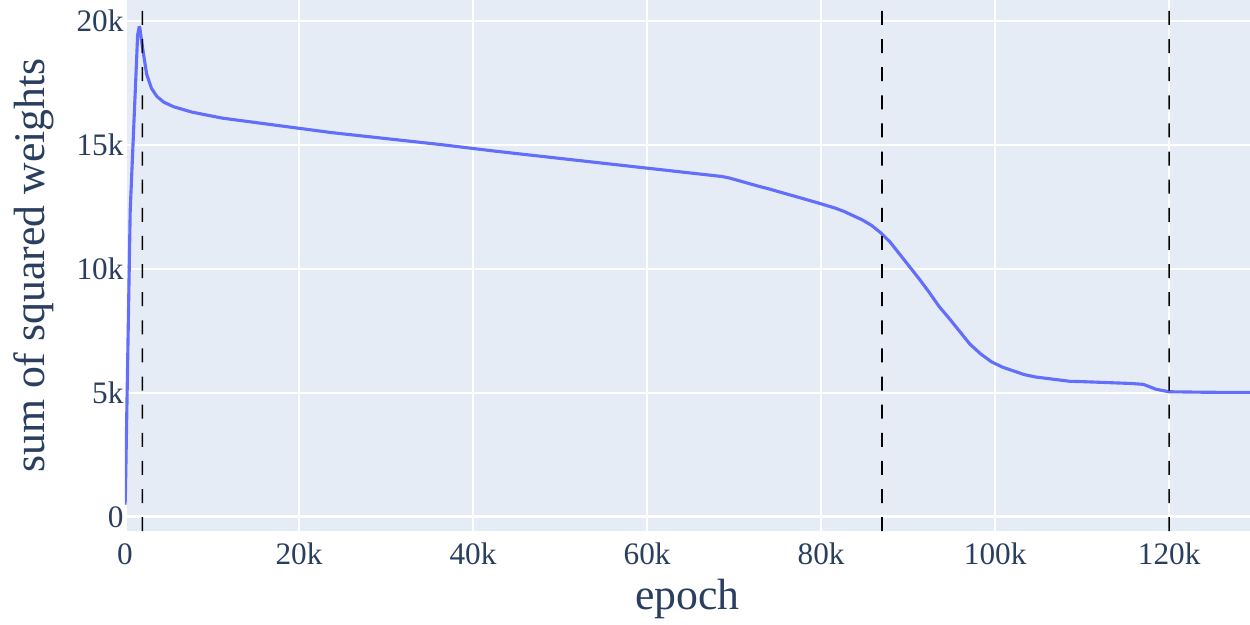}
    \end{subfigure}
    \begin{subfigure}{}
    \includegraphics[width=0.49\textwidth]{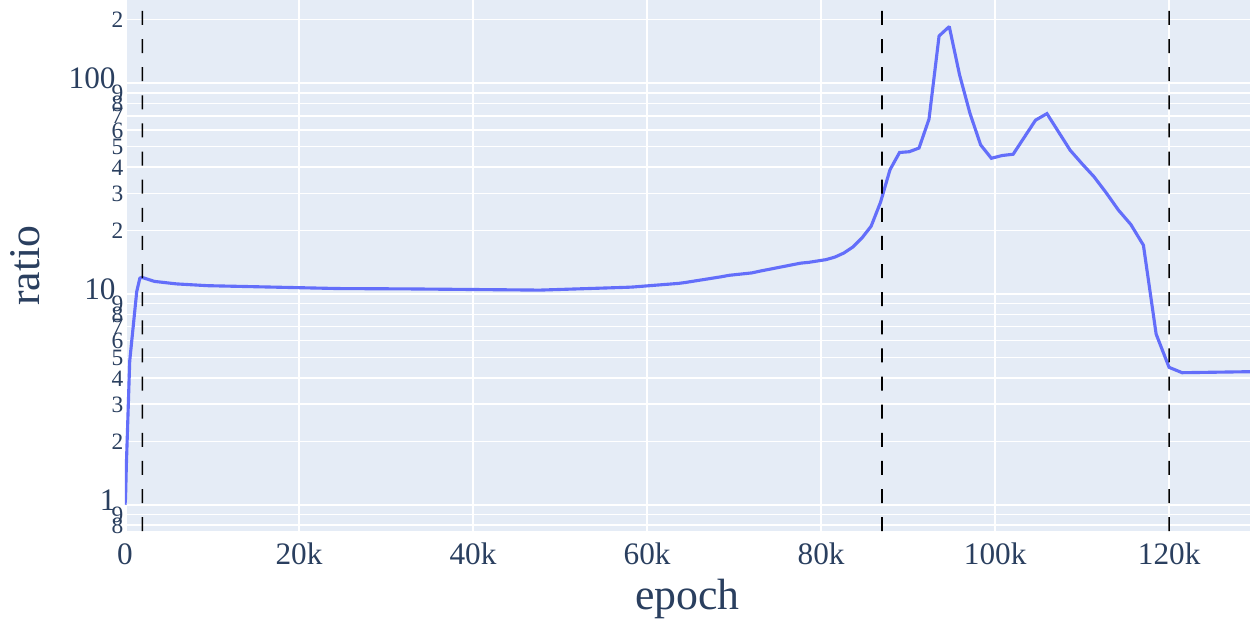}
    \end{subfigure}
    \caption{The sum of squared weights (\textbf{left}), and ratio of test loss and restricted loss (\textbf{right}). The sum of squared weights decreases smoothly during circuit formation and more sharply during cleanup, indicating both phases are linked to weight decay. Intuitively, restricted loss is us artificially cleaning up some the model (besides $W_U$), while test loss requires both circuit formation and cleanup. So a large discrepancy shows the rate of circuit formation outstrips the rate of cleanup during grokking.}
    \label{fig:further_progress_measures}
\end{figure}

\begin{figure}[!ht]
     \centering
     \begin{subfigure}{}
         \includegraphics[width=0.49\textwidth]{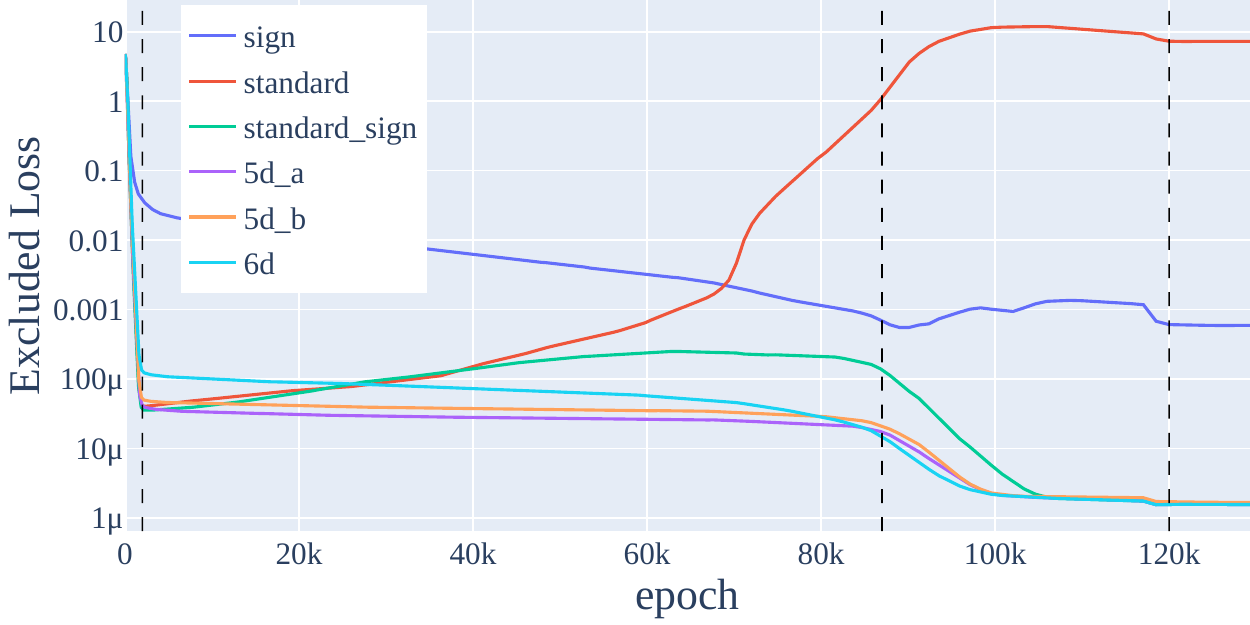}
     \end{subfigure}
     \begin{subfigure}{}
         \centering
         \includegraphics[width=0.49\textwidth]{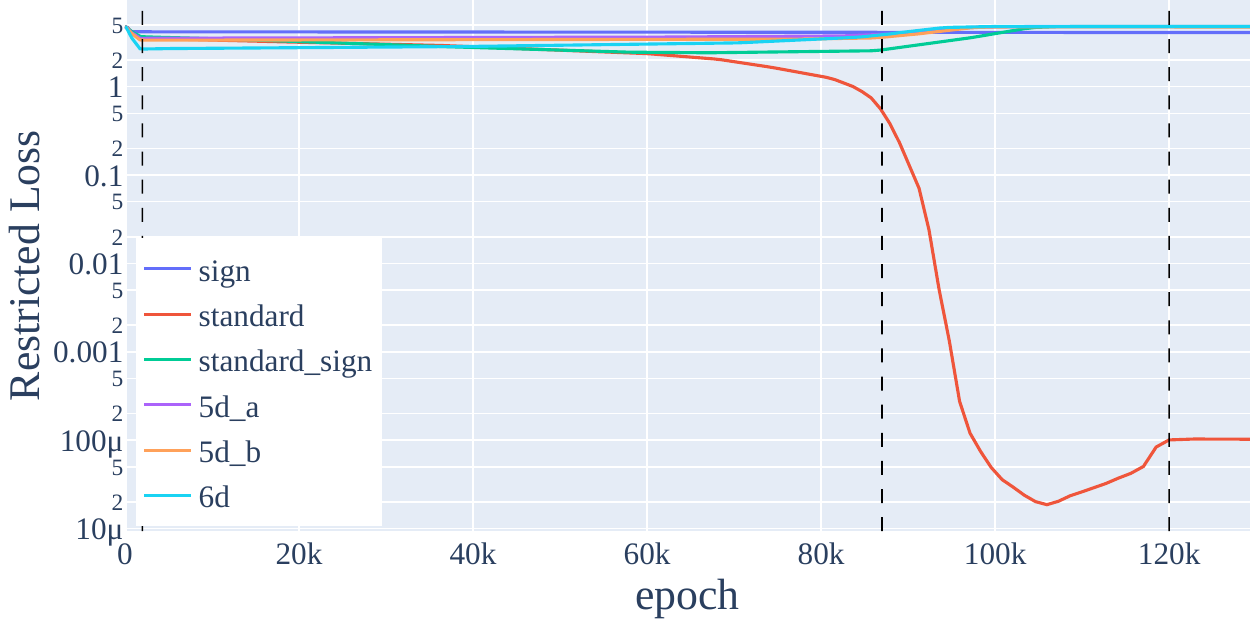}
     \end{subfigure}
     \caption{Excluded (\textbf{left}) and restricted loss (\textbf{right}), separated out by representation. As with the results of Section~\ref{sec:progress_measures}, this shows the model interpolates between memorizing and generalizing. In the restricted loss plot, we see the sign representation is incapable of solving the task alone, but contributes several orders of loss improvement when coupled with the standard representation, as can be seen in excluded loss.}
\end{figure}

Here, we provide further discussion on how we use progress measures to understand grokking generalization in our models. We first give more full definitions of our progress measures below.

\textbf{Restricted Loss.} We restrict the MLP activations to the terms corresponding to $\rho(ab)$ in the key representations, a $16 + 1$ dimensional subspace of $\R^{128}$, and then map this restricted MLP layer to logits. By doing so, we isolate the performance of the generalising algorithm. This assumes that the memorising algorithm has no privileged subspace in the MLP layer.

\textbf{Excluded Loss.} The opposite of restricted loss. Instead of keeping the key representations, we remove only those representations from the MLP neurons, and see how this affects loss. Having removed the generalising solution, this isolates the performance of the memorising solution. This therefore makes sense to measure only on the \emph{training} data, which we do.

The three phases of training we define are as follows, and can be seen in Figures~\ref{fig:progress_measures} and \ref{fig:further_progress_measures}.

\textbf{Memorization.} (Epochs 0-2k) We first observe a decline of both excluded and train loss, with test and restricted loss both remaining high. In other words, the model memorizes the training data. The sum of squared weights peaks at the end of memorization, so weight decay does not prefer these memorized circuits. As test loss increases but restricted loss stays constant as no progress towards generalization is made, the ratio of test loss to restricted loss rises.

\textbf{Circuit Formation.} (Epochs 2.2k-87k) In this phase, excluded loss rises, sum of squared weights falls (Figure~\ref{fig:further_progress_measures}), restricted loss starts to fall, and train and test loss stay flat. This suggests that the models behavior on the train set transitions smoothly from the memorising solution to the generalizing solution. The fall in the sum of squared weights suggests that circuit formation likely happens due to weight decay. Notably, the circuit is formed well before grokking.

\textbf{Cleanup.} (Epochs 87k-120k) In this phase, restricted loss continues to drop, test loss suddenly drops, sum of squared weights sharply drops, and the ratio of test to restricted loss is variable and then sharply decreases (Figure~\ref{fig:further_progress_measures}). As the generalising circuit both solves the task well and has lower weight at comparable performance as compared with memorisation circuits on the training set, weight decay encourages the network to shed the memorised solution. Weight decay contributes an important inductive bias of our networks \cite{neyshaburSearchRealInductive2015}. The slight rise in restricted loss at the very end of training is too a result of weight being traded off against performance -- multiplying the entire circuit by a fixed constant $r > 1$ will reduce loss, though also requires more weight.

\begin{figure}[t]
     \centering
     \begin{subfigure}{}
         \includegraphics[width=0.49\textwidth]{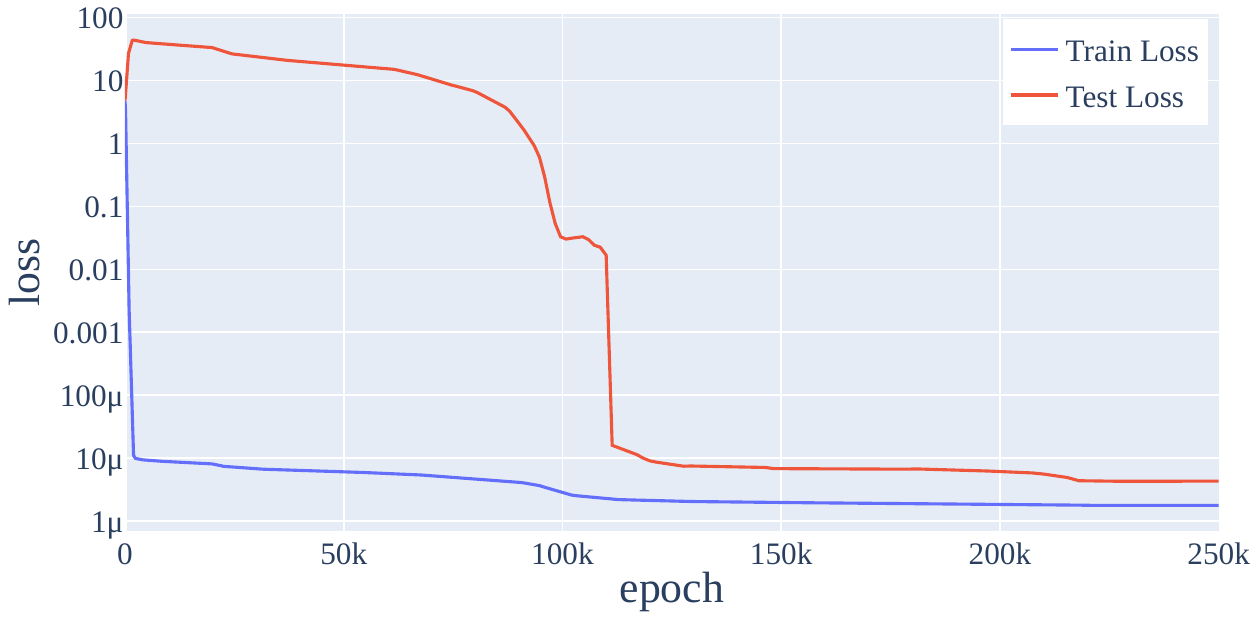}
     \end{subfigure}
     \begin{subfigure}{}
         \centering
         \includegraphics[width=0.49\textwidth]{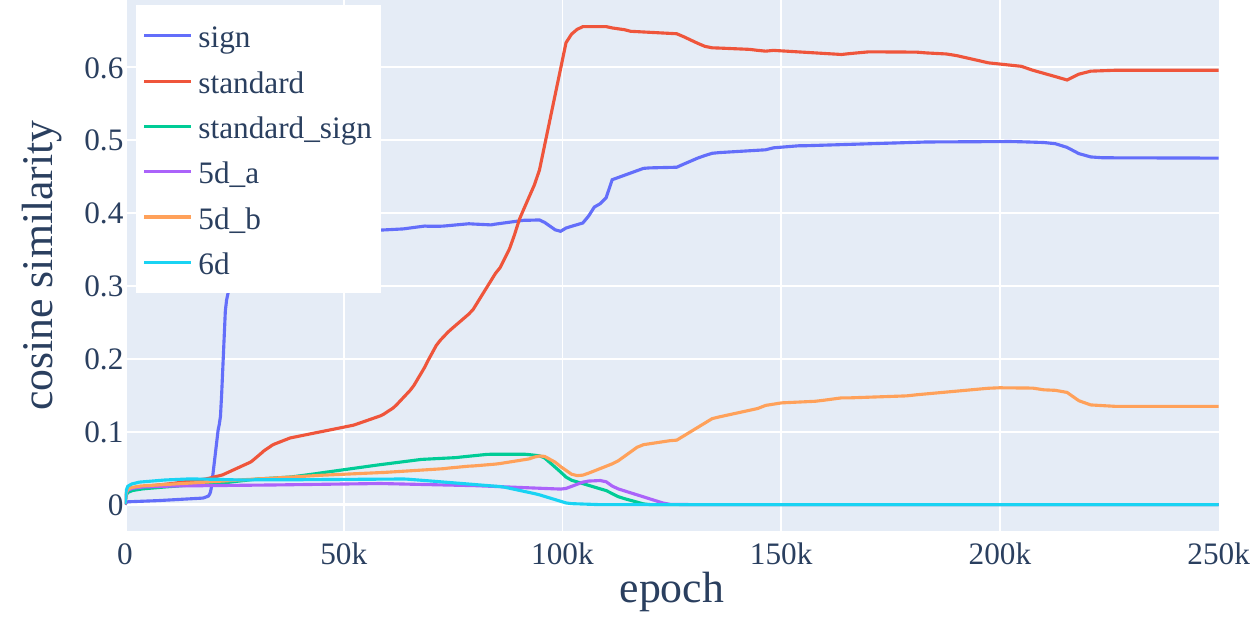}
     \end{subfigure}
     \caption{(\textbf{Left}) Train and test loss of the mainline model, only on a different random seed. (\textbf{Right}) Logit similarity of this run over training. We see two phases of grokking. The model initially groks as the memorizing circuit is cleaned up in presence of the valid general standard circuit. Loss then plateaus as the 5d\_b circuit is learned around epoch 100k, before the model groks again as cleanup continues.}
     \label{fig:phase_change_grok}
\end{figure}

\subsection{Full Circuit Analysis: Sign Representation}
\label{sec:sign_circuit}

In \citet{nandaProgressMeasuresGrokking2023}, the authors primarily analyze 2d representations via Fourier transforms, and we primarily analyze 4d standard representations in our mainline model. Treating sines and cosines as separate objects adds complexity, which we avoid by unifying them as matrix elements of the same representation. However, two dimensional features retain some redundancy over choice of basis, or equivalently, choice of rotation axis. So in general, some manipulation of activations and weights is necessary to interpret the model.

The sign representation on the other hand is a one dimensional representation of certain groups. This computational subgraph may be understood by directly inspecting activations and weights, without ever having to change basis. We demonstrate this simplicity on our mainline model.

\textbf{MLP neuron activations are `blocky'}. We can identify interpretable activation patterns by inspection. Working backwards we identify embeddings directly learn $\pm sign(a)$ and $\pm sign(b)$. 

\begin{figure}[!ht]
    \centering
    \includegraphics[width=0.7\columnwidth]{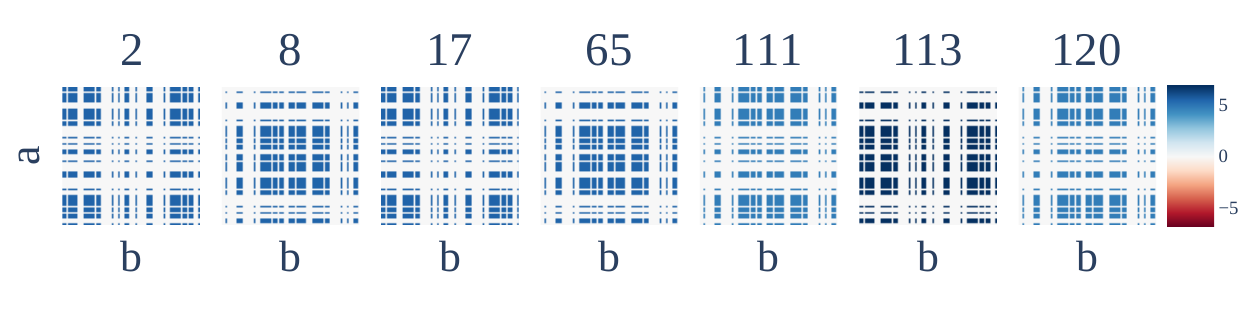}
    \caption{The seven `sign neuron' activations over the whole distribution of inputs. Each activates uniformly on inputs, with form some multiple of $\mathbbm{1}(sign(a)=\pm 1)\mathbbm{1}(sign(b)=\pm 1)$, where $\pm$ are independent.}
    \label{fig:blocky_sign_neurons}
\end{figure}

We then can write out, for $x,y$ some positive constants and $n_i$ neuron $i$:
\begin{align*}
n_{2} &= x ReLU \left(+ sign(a) + sign(b)\right)\\
n_{8} &= x ReLU \left(- sign(a) - sign(b)\right) \\
n_{17} &= x ReLU \left(+ sign(a) + sign(b)\right) \\
n_{65} &= x ReLU \left(- sign(a) - sign(b)\right) \\
n_{111} &= x ReLU \left(+ sign(a) - sign(b)\right) \\
n_{113} &= y ReLU \left(- sign(a) + sign(b)\right) \\
n_{120} &= x ReLU \left(+ sign(a) - sign(b)\right) \\
\end{align*}
In general, interpreting the matrix multiplication operation is challenging, though in the one dimensional  case it turns out to be simple. We see that the MLP performs multiplication of signs via ReLU and addition. For instance

\begin{equation*}
n_2 + n_8 + n_{111} + n_{113} = 2x \times sign(a) \times sign(b)
\end{equation*}

This is essentially computing an XOR gate on the inputs, and in particular not multiplication of arbitrary inputs, which is why the network can implement this operation perfectly. Note that we need a minimum of four neurons to implement this operation in this manner \footnote{If $x, y \in \{0, 1\}$ then $x$ XOR $y = ReLU(x-y) + ReLU(y-x)$ is a solution in two neurons.}. Empirically, we found that the number of sign neurons was often four exactly. In this case, neuron 113 appears to be used in two such multiplication calculations.

We expect that higher dimensional matrix multiplication is implemented similarly -- see further discussion in Appendix~\ref{app:matmuls}.

\textbf{Map to logits}. Calling neurons 2, 8, 17 and 65 positive, and neurons 111, 113, and 120 negative, we find that $W_U|_+ \sim + sign(c^{-1})$ and $W_U|_- \sim -sign(c^{-1})$, thus this circuit contributes positively to logits on correct signs and negatively to wrong signs, giving a contribution $\chi_{sign}(abc^{-1})$ to logits.

\subsection{Implementing Multiplication via ReLUs}
\label{app:matmuls}

Here we briefly discuss how networks may implement multiplication in a single layer. Our \abbrevalgorithm{} algorithm necessitates this in step 2, and we provide a simple example of this occurring in Appendix~\ref{sec:sign_circuit}.

Networks can multiply activations to some extent in one layer, though may not be able to do so perfectly, and also may put redundant information into additional directions (as we suspect comprises the 12\% residual of standard MLP neurons in Section~\ref{sec:mlp_neurons}). Note in this context that multiplication is not generic multiplication, but multiplication of a fixed set of elements. Most of our representation matrices have entries $\{0, -1, 1\}$\, on which multiplication can be implemented in a finite set of ReLU's with a bias as for instance

$$ x \times y = ReLU (x+y-1) + ReLU (-x-y-1) - ReLU (x-y-1) - ReLU (-x+y-1)$$

Changing the network architecture may aid it's ability to perform multiplication. Changing activation function to $x^2$ for instance permits multiplication generically as

$$ x \times y = \frac{1}{4}\left((x+y)^2 - (x-y)^2\right)$$

We hypothesize that the number of neurons in each representation cluster learned is linked to the number of such ReLU activations required to compute the matrix multiply, explaining why we have many more standard neurons than sign even after accounting for higher dimensionality of representation. Of course networks won't implement elementwise multiplication but rather some efficient matrix algorithm, such as Strassens.

\subsection{Visualizing the Embeddings and Unembeddings.}

\begin{figure}[h]
     \centering
     \begin{subfigure}{}
         \includegraphics[width=0.49\textwidth]{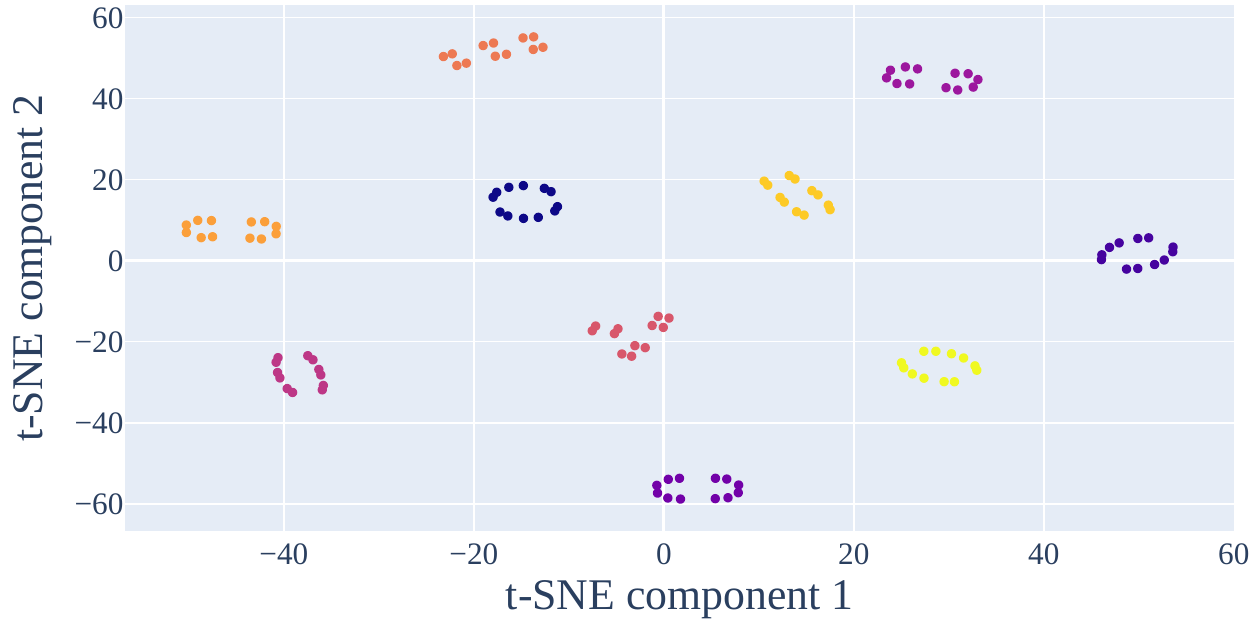}
     \end{subfigure}
     \begin{subfigure}{}
         \centering
         \includegraphics[width=0.49\textwidth]{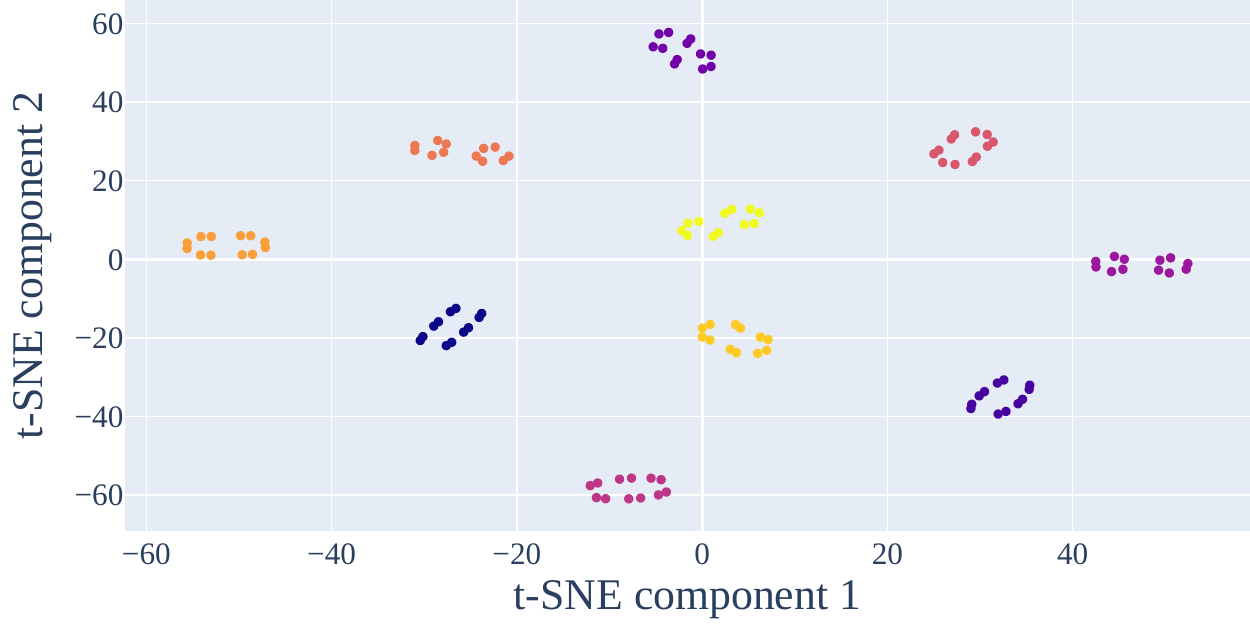}
     \end{subfigure}
     \begin{subfigure}{}
         \centering
         \includegraphics[width=0.49\textwidth]{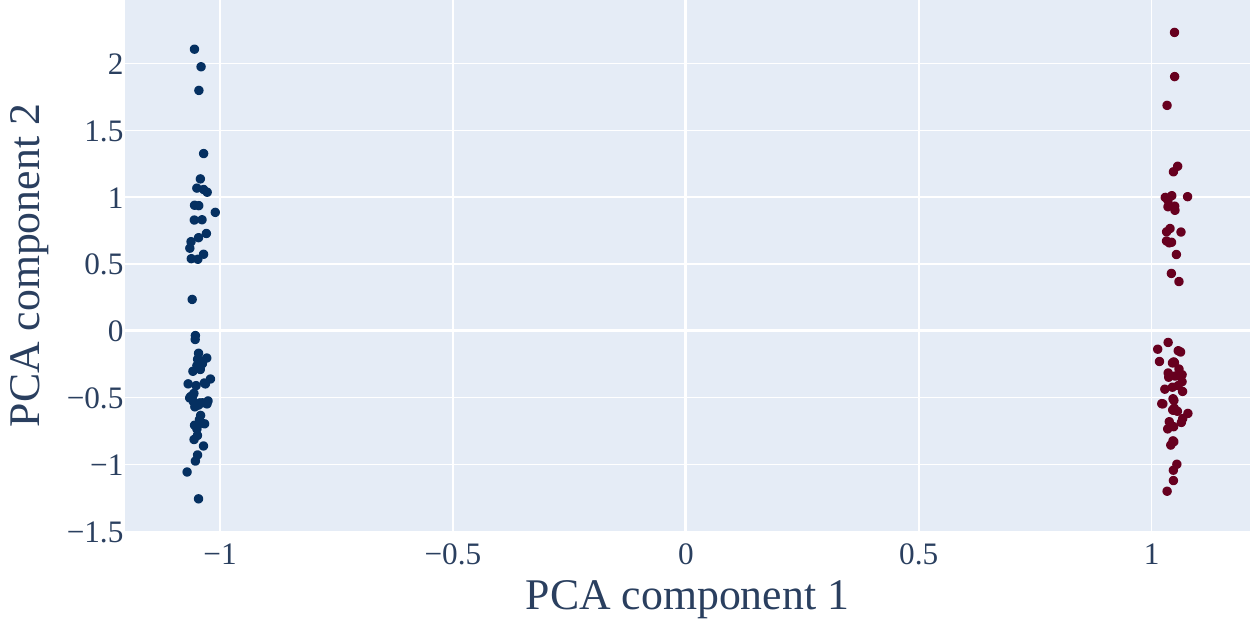}
     \end{subfigure}
     \caption{Left embedding (left) and right embedding (right) visualized in two dimensions via t-SNE. We see a large amount of structure. Clusters correspond to cosets of an order 12 subgroup of $S_5$. (Bottom) Visualization of the unembedding via PCA. The two clusters correspond to cosets of $A_5$, the alternating group i.e. the sign of group elements.}
     \label{fig:visualisations}
\end{figure}

\citet{powerGrokkingGeneralizationOverfitting2022} found it useful to use t-SNE to vizualize the unembedding in their networks trained on $S_5$. Here, we replicate their results, and show an additional meaningful visualization in Figure~\ref{fig:visualisations}. We did not find the unembedding $W_U$ to cluster into subgroups as they did via t-SNE, but did via PCA. We did find clustering in embeddings into cosets of a subgroup of $S_5$, though found the cosets to be of a different subgroup. Over different runs, these subgroups were arbitrary, though given the stochastic nature of t-SNE it is hard to say whether this observation is meaningful. 

\subsection{Further Reverse Engineering Details}
\label{sec:gory-details}

\textbf{Logit Similarity.} Observed logits $l(a,b,c)$ are an $n^3$ dimensional tensor over all input pairs $(a,b)$ and outputs $c$. The  \abbrevalgorithm{} algorithm's character predictions $\chi(abc^{-1})$ are also an $n^3$ tensor. We compute the correlation of these by flattening each tensor into a vector of dimension $n^3$, and computing the cosine similarity of these.

\textbf{Representation Space and Projection.} We perform an operation analogous to extracting the Fourier modes of a periodic function at each frequency. \footnote{The Fourier transform of a $\mathbb{C}$ valued function over $G$ can in fact be defined rigorously via representation theory. We omit details here.} Each representation gives a set of $n$ $d \times d$ matrices, one for each group element. We wish to investigate to what degree these are present in various model weights or activations. We can think of each representation as an  $n \times d^2$ tensor of flattened matrices $R$. We call the $n$-dimensional space spanned by these $d^2$ columns \emph{representation space}.  In order to project onto this space, we apply QR decomposition to $R$, obtaining $\tilde{R}$.

Any embedding or unembedding can be thought of an an $n \times h$ tensor $W$. Then $\tilde{R}^T W$ is a $d^2 \times h$ matrix. By inspecting the $h$ dimension of this matrix, we may understand neuron clustering, and by comparing the norm of it relative to the norm of the embedding or unembedding, we understand the percentage contribution of the subspace. An entirely analogous methodology is applied to understanding the MLP neurons via hidden representation spaces.

\textbf{Centering.} Neural network activations often contain large biases, even without the presence of explicit bias terms in the architecture. MLP neurons follow a ReLU activation, so necessarily have a mean positive activation. Accounting for this would artificially increase all `fraction of variance explained' metrics. To avoid this, we remove this bias by subtracting the mean over the batch dimension before interpreting the MLP activations. 

Similarly, since softmax is a function of relative logit difference, on each fixed input logits have some learned and unimportant bias. Accounting for this would artificially contribute to `logit similarity' under the trivial representation, and artificially increase the fraction of logits explained metric. To avoid this, we remove this by subtracting the mean over output dimension.

\section{Further Future Work}
\label{app:future_work}

Below, we outline an additional area of future work.

\textbf{Further group theoretic tasks.} In this work we focus on the task of group composition. \citet{powerGrokkingGeneralizationOverfitting2022} find that several other binary operations on pairs of input elements also grok, some of which are valid on any group. A trivial extension would be to the task $(a,b) \to ab^{-1}$, which may be solved simply by learning a permutation of the right embedding. A non trivial extension would be to conjugacy $(a, b) \to aba^{-1}$, which is of mathematical significance. Or to $(a,b,c)\to abc$. Each of these may be solved via similar representation theoretic algorithms, though we hypothesize would require two ReLU layers to implement two matrix multiplies. Other classes of group theoretic tasks include those of group actions (a superset of group composition type tasks) or to group theoretic automata, where we expect representation theoretic algorithms to apply too. Extending to semigroups (arbitrary associative multiplication tables) expands the set of tasks one could model, though there is no equivalent of representation theory for semigroups.

%\citet{hiltonUnderstandingRLVision2020} propose the diversity hypothesis through understanding vision in RL: \textit{Interpretable features tend to arise (at a given level of abstraction) if and only if the training distribution is diverse enough (at that level of abstraction).} This suggests that studying universality requires enough variation in the training distribution, which our group theoretic tasks do not contain here. ...

\section{Further Discussion on Inductive Biases}
\label{sec:inductive_bias}

A key question in ML is of understanding the inductive biases of a network: what are the class of algorithms natural for a network to express? In addition to being useful across the board, results in this area could help guide hypothesis formation in mechanistic interpretability. Examples of preliminary work on understanding Transformer inductive biases is presented in \citet{weissThinkingTransformers2021} and \citet{lindnerTracrCompiledTransformers2023}. 

Our work is useful in demonstrating the importance of \textit{linearity} in networks. At a first glance, our algorithm seems an overly complex solution to the problem to us, requiring some advanced mathematics to understand. Yet, networks are extremely good at multiplying vectors of activations by matrices of parameters. Our algorithm consists mostly of these operations, with a single step of activation-activation multiplication in the middle to implement the matrix multiply $\rho(a), \rho(b) \to \rho(a)\rho(b)$. We discuss how networks may implement this operation in Appendix~\ref{app:matmuls}. Note that the distinction between parameter-activation multiplication and activation-activation is important, with the former substantially easier for networks to implement. We also use a factored architecture (Appendix~\ref{sec:architecture_details}), which results in a low-rank implicit bias, which may encourage a sparse number of representations to be learned. More subjectively, we found the process of reasoning through the algorithm and its implementation in the model to be insightful for better understanding networks ourselves. 

We view this as evidence that the class of functions natural to humans and natural to networks are fundamentally different. Gaining examples like these is a step forward, but much more future work remains to be done in gaining a better understanding on these topics. Findings like these have in the past been beneficial in understanding real behaviour in networks, for instance the induction heads found by \citet{olssonIncontextLearningInduction2022} are an important part of the circuit for indirect object identification found by \citet{wangInterpretabilityWildCircuit2022}.

\section{Universality Results}
\label{sec:univ_results}

Here, we give full, unaveraged summary statistics of our runs on 4 seeds discussed in Section~\ref{sec:universality}. We omit the 50 MLP $S_5$ runs.

\textbf{Why are logit FVE scores relatively low?}
We note that our algorithm's prediction does not explain all of the logits, as can be seen from FVE being less than 100\%. Here, we provide some discussion on this. We do not believe this is evidence against our claim that we completely understand the important algorithms our network is implementing. Rather, we believe this is a side product of limitations of our architecture. In particular, as we discuss in Section~\ref{sec:mlp_neurons} and Appendix~\ref{app:matmuls}, the model uses the ReLU activations to implement matrix multiplication of $\rho(a)$ and $\rho(b)$ to $\rho(ab)$. This can only be approximated by ReLUs, and produces other terms, notably significant components of $\rho(a)$ and $\rho(b)$. This means that the network cannot perfectly extract the components of $\rho(ab)$ because they do not correspond to directions orthogonal to all other terms, resulting in the map to logits $W_U$ extracting other terms and non-character terms being present in the logits. We suspect the high FVE of models trained on composition in the cyclic group is related to orthogonality properties of the discrete Fourier transform not shared by general irreducible representations.

\newpage 

\begin{table}[!ht]
\label{tb:mlp_univ}
\caption{Results from MLP runs on various groups and seeds. Our algorithm is universally learned. Key representations are listed in order learned.}
\vspace{0.5cm}
\tiny
    \centering
\input{data/mlp_all.tex}
\end{table}

\begin{table}[!ht]
\label{tb:transformer_uni}
\caption{Results from Transformer runs on various groups and seeds. Our algorithm is universally learned. Key representations are listed in order learned.}
\vspace{0.5cm}
\tiny
    \centering
    \input{data/transformer_all.tex}
\end{table}

\end{document}

%% file: data/all_avg.tex
\begin{tabular}{cccccccccccccccc} 
\toprule
\multicolumn{1}{l}{} & \multicolumn{8}{c}{\textbf{MLP}}                                                   & \multicolumn{7}{c}{\textbf{Transformer}}                                       \\
\cmidrule(r){2-9} \cmidrule(l){10-16}
\multicolumn{1}{l}{} & \multicolumn{5}{c}{FVE}                              & \multicolumn{3}{c}{Loss}    & 
\multicolumn{4}{c}{FVE}                        & \multicolumn{3}{c}{Loss}      \\
\cmidrule(r){2-6} \cmidrule(lr){7-9}\cmidrule(lr){10-13} \cmidrule(r){14-16}
\textit{Group}       & \textit{$W_a$~} & \textit{$W_b$~} & \textit{$W_U$~} & \textit{MLP~} & \textit{$\rho(ab)$~} & \textit{Test~} & \textit{Exc.~} & \textit{Res.} & $W_E$   & \textit{$W_L$~} & \textit{MLP~} & \textit{$\rho(ab)$~} & \textit{Test} & \textit{Exc.} & \textit{Res. }  \\
\midrule
$C_{113}$              & 99.53\%  & 99.39\% & 98.05\% & 90.25\% & 12.03\%     & 1.63e-05 & 5.95  & 6.88e-03 & 95.18\% & 99.52\%  & 92.12\% & 16.77\%         & 2.67e-07 & 9.42   & 2.12e-02  \\
$C_{118}$              & 99.75\%  & 99.74\% & 98.43\% & 95.84\% & 13.26\%     & 5.39e-06 & 8.72  & 3.60e-03 & 94.05\% & 99.64\%  & 94.63\% & 17.11\%         & 1.73e-07 & 15.93  & 2.55e-01  \\
$D_{59}$               & 99.71\%  & 99.73\% & 98.52\% & 87.68\% & 12.44\%     & 6.34e-06 & 12.37 & 1.60e-06 & 98.58\% & 98.53\%  & 85.01\% & 10.85\%         & 3.20e-06 & 46.42  & 2.82e-05  \\
$D_{61}$               & 99.26\%  & 99.45\% & 98.26\% & 87.61\% & 12.48\%     & 1.79e-05 & 12.00 & 1.69e-06 & 98.33\% & 97.40\%  & 85.59\% & 11.11\%         & 1.63e-02 & 41.64  & 9.60e-02  \\
$S_{5}$                & 100.00\% & 99.99\% & 94.14\% & 88.91\% & 12.13\%     & 1.02e-05 & 11.72 & 2.21e-07 & 99.84\% & 99.97\%  & 85.28\% & 10.23\%         & 1.43e-07 & 17.77  & 4.44e-09  \\
$S_{6}$ &    99.65\% &    99.78\% &    93.67\% &  86.38\% &              8.98\% &   4.95e-05 &      12.17 &   2.66e-06 &    99.94\% &    99.93\% &  86.32\% &              9.35\% &   2.21e-06 &     291.67 &   1.05e-06 \\
$A_5$                & 99.04\%  & 99.31\% & 93.27\% & 86.69\% & 10.26\%     & 1.94e-05 & 9.82  & 5.28e-07 & 97.53\% & 97.40\%  & 83.56\% & 8.22\%          & 4.88e-02 & 19.76  & 7.70e-04  \\
\bottomrule
\end{tabular}

%% file: data/mlp_all.tex
\begin{tabular}{cccccccccccc} 
\toprule
\multicolumn{1}{l}{} & \multicolumn{1}{l}{} & \multicolumn{1}{l}{}                                   & \multicolumn{3}{c}{Loss}    & \multicolumn{6}{c}{FVE}                                         \\
\cmidrule(r){4-6} \cmidrule(l){7-12}
Group                & Seed                 & Key Representations                                             & Test     & Exc.  & Res.     & Logit   & $W_a$    & $W_b$    & $W_U$   & MLP     & $\rho(ab)$  \\ 
\midrule
$C_{113}$              & 1                    & 32, 50, 44, 16, 17, 34, 4, 11, 22, 1, 8, 13, 25        & 2.86e-05 & 6.22  & 9.84e-03 & 93.68\% & 99.23\%  & 99.07\%  & 97.95\% & 91.08\% & 11.80\%     \\
$C_{113}$              & 2                    & 7, 22, 25, 24, 36, 30, 14, 41, 44, 48, 50              & 5.89e-06 & 5.15  & 1.96e-06 & 98.08\% & 99.65\%  & 99.81\%  & 98.35\% & 89.95\% & 11.71\%     \\
$C_{113}$              & 3                    & 37, 14, 55, 47, 52, 34, 9, 5, 54, 45, 2, 3, 18, 28, 39 & 2.46e-05 & 5.69  & 8.29e-03 & 92.69\% & 99.57\%  & 98.96\%  & 97.85\% & 90.01\% & 12.14\%     \\
$C_{113}$              & 4                    & 55, 11, 30, 27, 43, 34, 29, 22, 3, 53                  & 6.15e-06 & 6.74  & 9.38e-03 & 97.89\% & 99.65\%  & 99.72\%  & 98.06\% & 89.98\% & 12.46\%     \\
$C_{118}$              & 1                    & 31, 49, 1, 12, 22, 45, 2, 20, 24, 28, 44, 56           & 5.75e-06 & 8.53  & 3.90e-06 & 97.22\% & 99.55\%  & 99.55\%  & 98.46\% & 94.61\% & 12.66\%     \\
$C_{118}$              & 2                    & 47, 23, 19, 29, 16, 44, sign, 30, 32, 38, 46, 58       & 5.19e-06 & 7.52  & 2.90e-03 & 98.25\% & 99.88\%  & 99.81\%  & 98.36\% & 94.67\% & 13.59\%     \\
$C_{118}$              & 3                    & 16, 30, 39, 8, 43, 48, sign, 11, 32, 40, 58            & 5.51e-06 & 10.92 & 7.93e-03 & 96.84\% & 99.84\%  & 99.77\%  & 98.40\% & 99.00\% & 13.63\%     \\
$C_{118}$              & 4                    & sign, 14, 5, 25, 57, 1, 22, 2, 4, 10, 28, 50           & 5.11e-06 & 7.90  & 3.56e-03 & 98.19\% & 99.73\%  & 99.82\%  & 98.49\% & 95.09\% & 13.17\%     \\
$D_{59}$               & 1                    & 19, 20, 15, 6, 14, 8, 3, 7, 12, 16, 21, 29             & 9.50e-06 & 9.17  & 7.41e-07 & 48.65\% & 99.46\%  & 99.40\%  & 98.58\% & 86.81\% & 11.65\%     \\
$D_{59}$               & 2                    & 18, 10, sign, 19, 6, 26, 7, 20, 21, 23                 & 4.30e-06 & 12.74 & 1.77e-06 & 54.92\% & 99.90\%  & 99.93\%  & 98.57\% & 88.05\% & 12.66\%     \\
$D_{59}$               & 3                    & sign, 20, 22, 16, 9, 12, 11, 15, 18, 19, 24            & 6.88e-06 & 11.06 & 1.91e-06 & 56.79\% & 99.57\%  & 99.71\%  & 98.50\% & 87.82\% & 13.05\%     \\
$D_{59}$               & 4                    & sign, 7, 10, 15, 21, 17, 19, 20, 29                    & 4.68e-06 & 16.52 & 1.98e-06 & 53.05\% & 99.90\%  & 99.89\%  & 98.42\% & 88.03\% & 12.41\%     \\
$D_{61}$               & 1                    & sign, 19, 23, 6, 7, 3, 24, 5, 12, 13, 14, 15           & 2.33e-05 & 11.21 & 1.76e-06 & 51.46\% & 99.17\%  & 99.63\%  & 98.05\% & 87.71\% & 12.05\%     \\
$D_{61}$               & 2                    & sign, 4, 29, 8, 27, 26, 19, 28, 14, 9, 2, 7, 3, 16, 18 & 2.87e-05 & 10.20 & 1.48e-06 & 53.04\% & 99.30\%  & 99.05\%  & 98.44\% & 87.15\% & 13.00\%     \\
$D_{61}$               & 3                    & 15, 14, 9, 26, 2, 25, sign, 28, 4, 18, 30              & 5.58e-06 & 14.86 & 1.74e-06 & 54.99\% & 99.68\%  & 99.89\%  & 98.28\% & 88.44\% & 12.60\%     \\
$D_{61}$               & 4                    & 20, 21, 19, 7, 17, 15, 23, sign, 14, 27, 30            & 1.39e-05 & 11.75 & 1.77e-06 & 50.33\% & 98.90\%  & 99.24\%  & 98.28\% & 87.15\% & 12.26\%     \\
$S_5$                & 1                    & sign, standard-sign, standard, 5d-a                    & 3.14e-05 & 10.09 & 1.52e-07 & 39.05\% & 100.00\% & 99.96\%  & 94.38\% & 87.95\% & 10.53\%     \\
$S_5$                & 2                    & sign, standard                                         & 2.94e-06 & 7.59  & 7.08e-07 & 84.81\% & 100.00\% & 100.00\% & 94.05\% & 88.88\% & 12.97\%     \\
$S_5$                & 3                    & sign, standard, 5d-b                                   & 4.32e-06 & 11.97 & 2.17e-08 & 59.89\% & 100.00\% & 99.99\%  & 94.97\% & 88.85\% & 12.38\%     \\
$S_5$                & 4                    & sign, standard                                         & 2.25e-06 & 17.21 & 1.96e-09 & 59.25\% & 100.00\% & 100.00\% & 93.18\% & 89.95\% & 12.66\%     \\
$S_6$                & 1                    & 5d-b, standard-sign, 5d-a, standard                    & 5.12e-05 & 12.97 & 1.98e-06 & 34.50\% & 99.77\%  & 99.87\%  & 93.25\% & 86.69\% & 8.38\%      \\
$S_6$                & 2                    & sign, standard, 5d-b                                   & 1.36e-05 & 13.42 & 2.52e-07 & 64.15\% & 100.00\% & 100.00\% & 93.42\% & 87.05\% & 10.27\%     \\
$S_6$                & 3                    & sign, standard-sign, 5d-a, 5d-b, standard              & 9.09e-05 & 10.86 & 6.87e-06 & 40.97\% & 98.96\%  & 99.42\%  & 94.42\% & 84.15\% & 7.52\%      \\
$S_6$                & 4                    & sign, 5d-b, standard-sign, standard                    & 4.21e-05 & 11.41 & 1.54e-06 & 56.96\% & 99.86\%  & 99.83\%  & 93.60\% & 87.64\% & 9.75\%      \\
$A_5$                & 1                    & standard, 3d-a, 3d-b                                   & 6.27e-05 & 7.23  & 1.56e-06 & 51.38\% & 98.52\%  & 98.69\%  & 93.08\% & 84.13\% & 9.46\%      \\
$A_5$                & 2                    & standard, 3d-a, 3d-b                                   & 5.09e-06 & 9.45  & 3.96e-07 & 43.86\% & 98.99\%  & 99.08\%  & 92.94\% & 85.11\% & 10.62\%     \\
$A_5$                & 3                    & 3d-a, 5d-a, standard                                   & 3.73e-06 & 11.55 & 5.70e-08 & 49.96\% & 99.53\%  & 99.74\%  & 92.73\% & 89.12\% & 10.81\%     \\
$A_5$                & 4                    & 5d-a, 3d-a, standard, 3d-b                             & 5.93e-06 & 11.03 & 9.81e-08 & 45.57\% & 99.14\%  & 99.73\%  & 94.32\% & 88.39\% & 10.14\%     \\
\bottomrule
\end{tabular}

%% file: data/transformer_all.tex
\begin{tabular}{ccccccccccc} 
\toprule
\multicolumn{1}{l}{} & \multicolumn{1}{l}{} & \multicolumn{1}{l}{}            & \multicolumn{3}{c}{Loss}     & \multicolumn{5}{c}{FVE}  \\
\cmidrule(r){4-6} \cmidrule(l){7-11}
Group                & Seed                 & Key Representations                      & Test     & Exc.   & Res.     & Logit   & $W_E$    & $W_U$    & MLP     & $\rho(ab)$  \\ 
\midrule
$C_{113}$              & 1                    & 16, 30, 56                      & 1.88e-07 & 9.77   & 2.26e-02 & 96.85\% & 90.08\%  & 99.49\%  & 92.67\% & 16.05\%     \\
$C_{113}$              & 2                    & 43, 53, 52, 49                  & 3.89e-07 & 8.45   & 1.45e-02 & 96.70\% & 96.91\%  & 99.71\%  & 89.72\% & 17.17\%     \\
$C_{113}$              & 3                    & 25, 56, 33, 19                  & 3.39e-07 & 8.76   & 8.19e-03 & 95.21\% & 95.70\%  & 99.23\%  & 93.32\% & 16.23\%     \\
$C_{113}$              & 4                    & 11, 12, 18                      & 1.53e-07 & 10.69  & 3.96e-02 & 97.93\% & 98.05\%  & 99.64\%  & 92.77\% & 17.62\%     \\
$C_{118}$              & 1                    & 37, 10, 16, sign, 19            & 1.67e-07 & 9.54   & 1.63e-03 & 98.55\% & 93.88\%  & 99.82\%  & 94.81\% & 17.54\%     \\
$C_{118}$              & 2                    & 8, 12, 27, 57                   & 1.94e-07 & 12.67  & 2.42e-03 & 98.35\% & 98.12\%  & 99.49\%  & 92.76\% & 16.36\%     \\
$C_{118}$              & 3                    & 53, 51, 4, 46                   & 1.74e-07 & 6.25   & 3.16e-03 & 98.59\% & 92.74\%  & 99.84\%  & 93.49\% & 14.48\%     \\
$C_{118}$              & 4                    & 17, sign, 29                    & 1.59e-07 & 35.28  & 1.01e+00 & 97.29\% & 91.46\%  & 99.42\%  & 97.46\% & 20.05\%     \\
$D_{59}$               & 1                    & sign, 21, 5, 2                  & 7.83e-06 & 54.66  & 1.06e-04 & 46.36\% & 98.36\%  & 95.46\%  & 85.38\% & 11.30\%     \\
$D_{59}$               & 2                    & 1, 15, 23                       & 3.76e-06 & 69.84  & 6.62e-08 & 51.28\% & 98.10\%  & 99.80\%  & 84.26\% & 10.00\%     \\
$D_{59}$               & 3                    & 22, 20, 26                      & 4.21e-07 & 31.95  & 6.76e-06 & 67.51\% & 99.12\%  & 99.36\%  & 85.09\% & 10.57\%     \\
$D_{59}$               & 4                    & 1, 16, sign, 24, 4              & 8.07e-07 & 29.24  & 1.23e-07 & 51.58\% & 98.76\%  & 99.49\%  & 85.31\% & 11.53\%     \\
$D_{61}$               & 1                    & 13, 26, 6, 16, 4, 1, 14, 12, 18 & 6.50e-02 & 27.57  & 3.41e-03 & 59.47\% & 95.48\%  & 95.39\%  & 86.08\% & 10.62\%     \\
$D_{61}$               & 2                    & sign, 24, 4, 18                 & 9.56e-06 & 51.42  & 3.80e-01 & 40.88\% & 98.91\%  & 94.59\%  & 85.70\% & 11.50\%     \\
$D_{61}$               & 3                    & 8, sign, 23, 28                 & 4.23e-07 & 53.44  & 6.87e-08 & 51.71\% & 99.06\%  & 99.82\%  & 85.38\% & 11.30\%     \\
$D_{61}$               & 4                    & 2, 6, 13                        & 1.89e-07 & 34.13  & 7.29e-08 & 56.37\% & 99.88\%  & 99.79\%  & 85.20\% & 11.04\%     \\
$S_5$                & 1                    & sign, standard-sign             & 1.42e-07 & 16.63  & 1.78e-09 & 62.85\% & 99.86\%  & 99.98\%  & 80.51\% & 8.44\%      \\
$S_5$                & 2                    & sign, standard-sign             & 2.41e-07 & 12.66  & 1.60e-08 & 73.46\% & 99.69\%  & 99.92\%  & 80.85\% & 7.58\%      \\
$S_5$                & 3                    & sign, standard                  & 9.39e-08 & 20.86  & 1.73e-11 & 59.06\% & 99.91\%  & 99.99\%  & 89.87\% & 12.44\%     \\
$S_5$                & 4                    & sign, standard                  & 9.51e-08 & 20.93  & 1.77e-11 & 59.07\% & 99.90\%  & 99.99\%  & 89.88\% & 12.46\%     \\
$S_6$                & 1                    & 5d-b                            & 2.26e-06 & 531.31 & 4.93e-16 & 44.50\% & 99.97\%  & 100.00\% & 88.06\% & 9.57\%      \\
$S_6$                & 2                    & sign, 5d-b                      & 2.91e-06 & 62.67  & 4.19e-06 & 65.69\% & 99.86\%  & 99.74\%  & 80.58\% & 7.90\%      \\
$S_6$                & 3                    & sign, 5d-b                      & 1.82e-06 & 286.64 & 9.78e-12 & 49.53\% & 99.96\%  & 100.00\% & 88.32\% & 9.96\%      \\
$S_6$                & 4                    & sign, 5d-b                      & 1.87e-06 & 286.08 & 1.02e-11 & 49.58\% & 99.96\%  & 100.00\% & 88.32\% & 9.96\%      \\
$A_5$                & 1                    & 3d-a, 3d-b                      & 1.35e-07 & 15.65  & 1.40e-03 & 63.19\% & 94.62\%  & 95.00\%  & 77.00\% & 6.30\%      \\
$A_5$                & 2                    & 3d-a, 3d-b                      & 1.30e-07 & 15.52  & 1.68e-03 & 63.29\% & 94.55\%  & 94.95\%  & 77.08\% & 6.33\%      \\
$A_5$                & 3                    & 5d-a, 3d-b, 3d-a                & 1.95e-01 & 27.92  & 1.30e-10 & 25.23\% & 101.04\% & 99.68\%  & 90.65\% & 8.52\%      \\
$A_5$                & 4                    & standard                        & 8.06e-08 & 19.95  & 1.21e-11 & 52.84\% & 99.92\%  & 99.98\%  & 89.52\% & 11.75\%     \\
\bottomrule
\end{tabular}